\documentclass[preprint,12pt]{elsarticle}

\usepackage[utf8]{inputenc} 
\usepackage[T1]{fontenc}    
\usepackage{hyperref}
\usepackage{url}            
\usepackage{booktabs}       
\usepackage{amsfonts}       
\usepackage{nicefrac}       
\usepackage{microtype}      

\usepackage{amsmath,amssymb}
\usepackage{amsthm}
\usepackage{bm}
\usepackage{array}          
\usepackage{mathptmx}       
\usepackage{tablefootnote}
\usepackage{mathtools}
\usepackage[mathcal]{eucal}
\usepackage{subcaption}
\usepackage{color}

\usepackage[frozencache,cachedir=.]{minted}

\newcommand{\todo}[1]{}
\renewcommand{\todo}[1]{{\color{red} TODO: {#1}}}

\usepackage{amssymb}

\renewcommand{\vec}[1]{\mathbf{#1}}
\renewcommand{\Re}{\mathbb{R}}

\DeclareMathOperator{\smape}{s\textsc{mape}}
\DeclareMathOperator{\mae}{\textsc{mae}}
\DeclareMathOperator{\mape}{\textsc{mape}}
\DeclareMathOperator{\mpe}{\textsc{mpe}}
\DeclareMathOperator{\ape}{\textsc{ape}}
\DeclareMathOperator{\pe}{\textsc{pe}}
\DeclareMathOperator{\pmape}{\textsc{p-mape}}

\DeclareMathOperator{\rmse}{\textsc{rmse}}

\DeclareMathOperator{\iqr}{\textsc{iqr}}

\DeclareMathOperator{\fc}{\textsc{FC}}
\DeclareMathOperator{\relu}{\textsc{ReLu}}

\newcommand{\nbeatsinput}{\vec{x}}
\newcommand{\nbeatshidden}{\vec{h}}
\newcommand{\nbeatsbackcast}{{\widehat{\nbeatsinput}}}
\newcommand{\nbeatsforecast}{{\widehat{\vec{y}}}} 
\newcommand{\windowlength}{w}

\journal{Applied Energy}

\begin{document}

\begin{frontmatter}



\title{N-BEATS neural network for mid-term electricity load forecasting}


\author[a1]{Boris N. Oreshkin}
\author[a2]{Grzegorz Dudek}
\author[a2]{Paweł Pełka}
\author[a3]{Ekaterina Turkina}

\address[a1]{Unity Technologies, 1751 Richardson Street, Suite 3.500, Montréal, QC H3K 1G6, e-mail: boris.oreshkin@gmail.com}
\address[a2]{Department of Electrical Engineering,
Czestochowa University of Technology, 42-200 Czestochowa, Al. Armii Krajowej 17, Poland, e-mail: dudek@el.pcz.czest.pl, p.pelka@el.pcz.czest.pl}
\address[a3]{HEC Montréal, 3000 Côte-Sainte-Catherine Road, Montréal, QC H3T 2A7, e-mail: ekaterina.turkina@hec.ca}

\begin{abstract}
This paper addresses the mid-term electricity load forecasting problem. Solving this problem is necessary for power system operation and planning as well as for negotiating forward contracts in deregulated energy markets. We show that our proposed deep neural network modeling approach based on the deep neural architecture is effective at solving the mid-term electricity load forecasting problem. Proposed neural network has high expressive power to solve non-linear stochastic forecasting problems with time series including trends, seasonality and significant random fluctuations. At the same time, it is simple to implement and train, it does not require signal preprocessing, and it is equipped with a forecast bias reduction mechanism. We compare our approach against ten baseline methods, including classical statistical methods, machine learning and hybrid approaches, on 35 monthly electricity demand time series for European countries. The empirical study shows that proposed neural network clearly outperforms all competitors in terms of both accuracy and forecast bias. Code is available here: \url{https://github.com/boreshkinai/nbeats-midterm}.

\end{abstract}







\end{frontmatter}


\section{Introduction}
\label{Intro}


Maintaining a continuous balance between electricity consumption and production is a prerequisite for power system stability and efficiency. At the same time, it poses a serious challenge, exacerbated in recent years by an increasing share of volatile, fluctuating renewable energy sources. The key requirement for balancing a power system is to have reliable forecasts of demand and generation at any time point. Accurate forecasts help avoid costs related to energy shortage or oversupply. For instance, a study from the California Energy Commission indicates improved solar and load forecasting can yield potential savings of USD 2 million yearly~\cite{Cal19}. Another report \cite{Hod15}, analyzing the California Independent System Operator market, shows that total cost savings from improved short-term wind power forecasting can be in the range of USD 5.05 to 146 million yearly depending on wind power. Therefore, accurate forecasts of electricity demand and supply are of great importance not only for ensuring safe and efficient system operation, but also for increasing market revenues and reducing financial risks.

Forecasting electricity demand at different horizons and granularity (hour, day, week, month, year, etc.) has traditionally been essential for supporting the operations of energy utility companies at different business levels. Vertically-integrated utilities used to rely on short-term forecasts (one hour to seven days) to mitigate energy supply interruption risks and long-term forecasts (one to twenty years) to plan future capacity investments. As the degree of deregulation and the ferocity of competition in electric power markets increased in the past decade, so did the need for more accurate mid-term load forecasts providing energy consumption forecasts one week to one year ahead. MTLF is necessary for power system operation and planning in such areas as maintenance scheduling, mid-term hydro thermal coordination, fuel reserve planning, energy import and export planning, revenue assessment for the utilities and security assessment~\citep{hammad2020methods}. Mid-term electricity load forecasting (MTLF) is crucial for negotiating forward contracts between generators and retailers or large consumers in deregulated power systems in the market for bilateral contracts, where the time frame for contracts can be several years \cite{And07}. Therefore, increasing MTLF accuracy translates directly into increased efficiency and safety, reduced financial risks and improved financial performance. The financial impact can be measured in millions of dollars for every point of forecasting accuracy gained. 

\subsection{Related Work}

\noindent\textbf{MTLF approaches} can be divided into two general categories~\cite{Ghi06}: conditional modeling and autonomous modeling. Conditional modeling focuses on the economic analysis, management and long-term planning and forecasting of energy load and energy policies. It uses exogenous variables that affect energy demand as the model inputs. For example, model inputs such as gross national product, consumer price index, exchange rates and average wage have been used in~\cite{Ghi06}. Additionally, variables describing network infrastructure and power system operation (e.g., length of transmission lines, number of highest voltage stations, number of connections, reserve margin and load diversity factor) can be introduced as additional inputs or factors guiding model selection~\cite{Moh18}. In autonomous modeling, the prediction of electricity demand is primarily based on historical demand, atmospheric temperatures, and variables expressing seasonality~\cite{Pei11}. This category is more appropriate for stable economies, with no sudden changes affecting electricity demand. 

The approach categories described above rely mostly on classical statistical and econometric modeling, although some approaches do use more advanced machine learning techniques~\cite{Sug11}. Classical approaches are largely based on variations of ARIMA, exponential smoothing (ETS) and linear regression. These models can deal with seasonal time series, but additional operations, such as decomposition, local approach or extension of the model with periodic components, are required~\cite{Dud16}. Classical models have inherent shortcomings related to limited adaptability and a shortage of expressive power to model non-linear relationships. This prompted researchers to take an interest in more flexible machine learning and computational intelligence models~\cite{Gon08}. Of these, neural networks (NNs) are the most explored in the field of forecasting. They have many attractive properties, such as the ability to model non-linear relationships and learn from data, universal approximation property and massive parallelism. Some examples of applying NN architectures to solve the MTLF problem include multilayer perceptrons (MLPs)~\cite{Chen17}, weighted evolving fuzzy NNs~\cite{Pei11} and NNs combined with linear regression and AdaBoost~\cite{Ahm19}.  

\noindent\textbf{Deep learning} (DL) has been very successful in solving complex forecasting problems in recent years. Its success can be largely attributed to increased model complexity and the ability to cross-learn on massive datasets including thousands and sometimes hundreds of thousands of time series. Thus, DL overcomes the fundamental limitations of classical NNs, such as the lack of expressive power and the inability to extract general patterns across multiple examples. Modern DL architectures are composed of combinations of basic structures, such as MLPs, recurrent NNs (RNNs) and convolutional NNs. A prominent problem in RNNs is vanishing or exploding gradient when processing long sequences. A long short-term memory network (LSTM) was proposed to solve it~\cite{Hoc97}. LSTM architecture is composed of a cell and several non-linear gates that control data flow inside the cell and decide on what information should be kept and what should be propagated to the next time step. LSTMs have been shown to outperform statistical and machine learning models such as ARIMA, support vector machine and classical NNs~\cite{Yan18}. In 2018, LSTM-based forecasting models won the M4 forecasting competition, which utilized 100,000 real-life time series~\cite{smyl2020hybrid}. Additionally, LSTM models have been widely used to solve load forecasting problems. For example, in the short-term forecasting context they were used alone~\cite{Nar17} and in combination with XGBoost~\cite{Zhe17}. In the mid-term load forecasting context, a pure LSTM approach was compared against a variety of ML based approaches~\cite{Dud20a}.       

In addition to the LSTM-based architectures mentioned above, the following DL architectures are promising in the context of forecasting \cite{Ben20}. First, WaveNet architecture was originally proposed for speech synthesis \cite{Oor16} and recently adapted to time series forecasting \cite{Bis19}. Unlike RNNs, which rely on sequential computation, WaveNet uses dilated causal convolutions, which are more efficient from a computation parallelism viewpoint and are advantageous while learning long-range dependencies. Second, encoder-decoder attention mechanism~\cite{Cho14} and Transformer~\cite{Vas17} offer another alternative to LSTMs. Attention helps to learn which parts of the input sequence are the most relevant to produce a correct prediction at the current time step. Unlike RNNs, attention-based architectures can focus on any part of the input sequence while producing a forecast, without being subject to the vanishing or exploding gradient problem. Transformer has demonstrated impressive forecasting accuracy results~\cite{Li19}. One of the downsides of transformer and of attention-based models in general is their high computational cost, which scales as a square of the input size. Finally, N-BEATS is a deep stack of fully connected layers connected with forward and backward residual links~\cite{Ore19}. N-BEATS has demonstrated state-of-the-art performance on multiple large-scale datasets, including M4's~\cite{Ore19} and is very computationally efficient (linear cost w.r.t. to input size).

\subsection{Motivation and Contributions}

In this paper, our main focus is mid-term electricity load forecasting with monthly granularity over a 12-month horizon. Existing work in the area of MTLF has a number of significant limitations and gaps, which we address and bridge in our current work. First, many of the existing works focus on linear or simple non-linear regression models that have limited modeling power~\citep{hammad2020methods}. 
In this research, we clearly demonstrate that advanced non-linear modeling relying on N-BEATS deep neural network architecture leads to substantial improvement in MTLF accuracy compared to such well-established models as ARIMA and ETS. Second, existing work exploring advanced forecasting techniques in MTLF has provided limited statistical analyses of forecasting errors, largely ignoring the analysis of the significance of forecasting accuracy gain (see for example the applications of machine learning~\cite{Ahm19}, bagging of classical models~\cite{DeOliveira18}, NNs~\cite{Chen17}, NNs in combination with kernel PCA~\citep{liu2019midterm} and soft computing~\cite{nti2019predicting}). In this paper, we clearly demonstrate that N-BEATS has a statistically significant forecasting accuracy gain with respect to statistical models as well as state-of-the-art machine learning and hybrid models. Our conclusions are based on the rigorous analysis of bootstrapped confidence intervals. Third, the original work on N-BEATS demonstrated its effectiveness on challenging competition datasets containing tens of thousands of time series from diverse domains~\cite{Ore19}. However, MTLF datasets tend to be rather small (dozens or hundreds of time series at best). This can be an obstacle for effectively applying powerful deep neural networks such as N-BEATS in the small-data regime. The downsides of operating in the small-data regime are three-fold. First, small dataset size exacerbates overfitting problems: large networks trained on few samples may have small error on the training set, but will fail to generalize on the unseen data in the test set. Second, hyperparameter values selected on a small validation set may be noisy and may fail to generalize on the test set. Finally, a common way to combat overfit is to reduce network capacity (e.g., reduce the width and the number of layers). There is significant uncertainty then whether the reduced-capacity network will be able to deliver the improved non-linear modeling capabilities to outperform classical statistical models on a small dataset. In this work, we clearly demonstrate that N-BEATS can effectively be applied to small-scale forecasting problems such as MTLF and result in state-of-the-art performance. Finally, existing forecasting algorithms focus only on improving forecasting accuracy and ignore the bias of the forecasts, which plays a crucial role in MTLF. Therefore, existing algorithms exhibit biased forecasts with uncontrollable forecasting bias when applied to the MTLF problem~\citep{Dud20a}. In this paper, we propose a simple and effective mechanism to control for forecasting bias using the pinball-$\mape$ loss function and demonstrate its effectiveness on real data.    

Our research contributions can be summarized as follows:
\begin{enumerate}
    \item We propose to apply the N-BEATS neural network in the context of MTLF and empirically demonstrate this powerful deep learning model's state-of-the-art performance in the small data regime. We make the code implementation of the method public under MIT license to facilitate the use of this method to solve difficult and challenging forecasting problems such as MTLF.
    \item We propose the pinball-$\mape$ loss function, which allows the model to directly minimize empirical $\mape$ loss while simultaneously controlling forecasting bias via a tunable hyperparameter. 
    \item We conduct a detailed empirical analysis of the forecasting results of 10 baseline algorithms and N-BEATS with proposed pinball-$\mape$ loss on a real dataset. Our analysis demonstrates statistically significant forecasting bias reduction and accuracy improvement of the proposed approach with respect to the 10 baselines including well-established statistical approaches and state-of-the-art domain-adjusted machine learning and hybrid approaches.
\end{enumerate}

The rest of the work is organized as follows. Section~\ref{N-BEATS} presents the proposed N-BEATS NN for MTLF. The experimental framework used to evaluate the proposed model's performance is described in Section~\ref{ER}. Finally, Section~\ref{Con} concludes the work.

\section{N-BEATS for MTLF} \label{N-BEATS}

\subsection{MTLF Task} 

We formulate the forecasting problem given a length $H$ forecast horizon and a length $T$ observed time series history $[y_1, \ldots, y_T] \in \Re^T$. The task is to predict the vector of future values $\vec{y} \in \Re^H = [y_{T+1}, y_{T+2}, \ldots, y_{T+H}]$ given past observations. For simplicity, we will later consider a \emph{lookback window} of length $w \le T$ ending with the last observed value $y_T$ to serve as model input, and we denote $\nbeatsinput \in \Re^w = [y_{T-w+1}, \ldots, y_T]$. We designate $\widehat{\vec{y}}$ the point forecast of $\vec{y}$. Its accuracy is evaluated using $\mape$, mean absolute percentage error~\citep{makridakis2000theM3},
\begin{align} \label{eqn:mape-definition}
\mape &= \frac{100}{H} \sum_{i=1}^H \frac{|y_{T+i} - \widehat{y}_{T+i}|}{|y_{T+i}|}.
\end{align}

\subsection{Lookback Window Selection} 

A few remarks are in order regarding the methodology for selecting the model input window size, $w$. The optimal choice of $w$ is very much problem dependent, and providing one specific value that will work for all problems is impossible. Therefore, $w$ is best treated as a hyperparameter whose optimal value is selected on a problem-specific validation set. At the same time, there exists a set of guidelines, which we present below, that usually help to significantly shrink the search space for $w$ given that we have access to some problem meta-data or results of exploratory analysis (such as presumed seasonality of the data, stochasticity of the data generating process, etc.). First, typically good values of $w$ are proportional to the seasonality period of the representative time series from the dataset. Rarely do we observe any useful information being extracted by the models when $w$ is a small fraction of the characteristic seasonality period of the problem. Second, if multiple seasonalities are present (e.g., 7 days and 24 hours), then the optimal $w$ may be a multiple of any of those seasonalities, depending on their mutual strength. Third, using an ensemble of several models, each trained on their own $w$, is usually extremely productive, and we observed great success applying this technique on many real-life proprietary datasets. For example, if we have monthly data with yearly seasonality, trying an ensemble with individual models trained with $w \in \{12, 24, 36, 48, 60\}$ may be a good idea. Fourth, smaller values of $w$ should be preferred for datasets containing short time series to avoid overfitting problems. Here, overfitting comes from two sources: (i) the increase in $w$ typically leads to a decrease in the number of training samples and (ii) for larger $w$, the overlapping training samples become more and more correlated, resulting in a decrease in the effective sample size, which exacerbates overfitting. Finally, if we anticipate that a given problem may have a swiftly changing generating process, small values of $w$ should be preferred, because historical information quickly gets outdated.

\begin{figure}[t]
\centering
\includegraphics[width=0.9\textwidth]{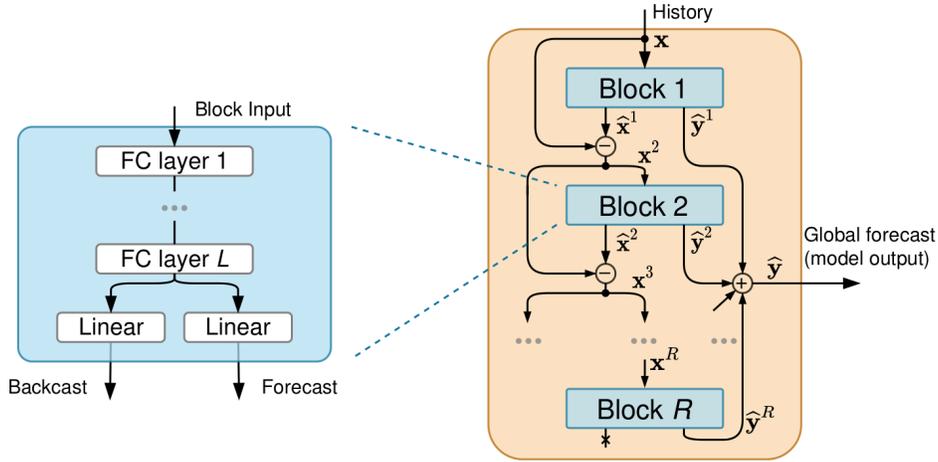}
\caption{N-BEATS block diagram.}
\label{fig:nbeats_architecture}
\end{figure}

\subsection{N-BEATS Architecture}

N-BEATS architecture is different from the existing architectures in a few aspects. First, instead of treating forecasting as a sequence-to-sequence problem, we treat it as a non-linear multivariate regression problem. Therefore, the basic building block of the architecture (see Fig.~\ref{fig:nbeats_architecture}, left) is a fully connected non-linear regressor that accepts the history of a time series and outputs multiple points in the forecasting horizon. Second, most existing time series architectures are relatively shallow (one to five LSTM layers, for example). We use the residual principle to stack many layers together (see Fig.~\ref{fig:nbeats_architecture}, right). For this, the basic block predicts both the future outputs and their contribution to the decomposition of the input, which we call backcast. It was demonstrated in~\cite{Ore19} that we can stack on the order of one hundred layers effectively using this principle, resulting in a very expressive model having very good generalization capabilities. Another advantage of the architecture is its simplicity that shows both at the conceptual and implementation levels. Conceptually, we can think about each fully connected layer as a multivariate linear regression block followed by a ReLu~\citep{nair2010rectified,Glorot+al-AI-2011-small} non-linearity. Therefore, N-BEATS can be thought of as simply being a multivariate regression that is repeated many times and interleaved with non-linearities. The conceptual simplicity translates into the implementation simplicity. The architecture can be coded in just 40 lines of code in standard TensorFlow~\cite{tensorflow2015whitepaper} syntax, as shown in the python code listing of the N-BEATS model presented in~Listing~\ref{listing:nbeats-inference-code} of~\ref{sec:nbeats-implementation}.

In terms of mathematical description, each block of N-BEATS is a sequence of fully connected layers with a forecast/backcast fork at the end. The architecture runs a residual recursion over the entire input window and sums block outputs to make its final forecast (see Fig.~\ref{fig:nbeats_architecture}). We assume that there are $R$ residual blocks each having $L$ hidden layers. If we refer, as is previously done, to $\nbeatsinput \in \mathbb{R}^{\windowlength}$ as the input of the architecture, use residual block and layer superscripts ($r$ and $\ell$, respectively) and denote the fully connected layer with weights $\vec{W}^{r,\ell}$ and biases $\vec{b}^{r,\ell}$ as $\fc_{r,\ell}(\nbeatshidden^{r, \ell-1}) \equiv \relu(\vec{W}^{r,\ell} \nbeatshidden^{r,\ell-1} + \vec{b}^{r,\ell})$, the operation of N-BEATS can be described as follows:
\begin{align}  \label{eqn:nbeats_fc_network}
\begin{split}
    \vec{x}^{r} &= \relu[\nbeatsinput^{r-1} - \nbeatsbackcast^{r-1}], \\
    \nbeatshidden^{r,1} &= \fc_{r,1}(\vec{x}^{r}), \  \ldots, \  \nbeatshidden^{r,L} = \fc_{r,L}(\nbeatshidden^{r,L-1}),  \\
    \nbeatsbackcast^{r} &= \vec{B}^{r} \nbeatshidden^{r,L}, \    \nbeatsforecast^{r} = \vec{F}^{r} \nbeatshidden^{r,L}.
\end{split}
\end{align}
We assume $\nbeatsbackcast^{0} \equiv \vec{0}$, $\nbeatsinput^{0} \equiv \nbeatsinput$, the projection matrices have dimensions $\vec{B}^{r} \in \mathbb{R}^{\windowlength \times d_h}$, $\vec{F}^{r} \in \mathbb{R}^{H \times d_h}$ and the final forecast is the sum of forecasts of all the residual blocks, $\nbeatsforecast = \sum_r \nbeatsforecast^{r}$.

\subsection{Pinball-$\mape$ Loss Function} \label{ssec:pinball_mape}

$\mape$ is a well-established performance metric for forecasting problems~\cite{makridakis2000theM3} and is the most commonly used accuracy measure in load forecasting. Training using $\mape$ as a loss function while $\mape$ is used for performance evaluation may be beneficial, because training and performance evaluation metric objectives are maximally aligned. Yet, this may result in forecasts that are biased, since forecast bias minimization is not directly instigated by $\mape$. To alleviate this problem, we propose pinball-$\mape$ ($\pmape$) evaluated over $N$ samples:
\begin{align}  \label{eqn:pinball_mape}
\pmape(y, \hat y) = \frac{1}{N} \sum_{i=1}^N
    \begin{cases}
         200 \cdot \tau (y_{i} - \hat y_{i}) / y_{i} & \text{if} \quad y_{i} \geq \hat y_{i} \\
         200 \cdot  (1-\tau) (\hat y_{i} - y_{i}) / y_{i}        & \text{otherwise}
    \end{cases}
\end{align}
The $\tau$ parameter in the $\pmape$ loss function can be adjusted on the validation set to compensate for biases arising from the training on $\mape$ loss. $\pmape$ loss with $\tau=0.5$ is equivalent to $\mape$ loss. Setting $\tau \in (0, 0.5)$ will tend to compensate for over-estimation bias, and setting $\tau \in (0.5, 1)$ will tend to compensate for under-estimation bias. We conjecture that a similar approach may be used with other loss functions ($\smape$, $\rmse$, etc.). Note that the use of lower $\tau$ values to avoid over-forecasting with $\mae$ based training was proposed by~\citet{smyl2020hybrid}. 

\section{Experimental Results} \label{ER}

In this section, we apply the proposed N-BEATS model to MTLF and compare its performance with that of other models based on classical statistical methods, machine learning methods and hybrid approaches.. 

\subsection{Dataset}

The models are applied to real-world data collected from \url{www.entsoe.eu} comprising monthly electricity demand for 35 European countries. The electricity demand time series in the dataset exhibit a non-linear trend, seasonality and a random component. The trend depends on the country's economic growth rate and climate change, such as global warming caused by greenhouse gas emissions and other factors~\cite{Hor05}. The seasonalities are related to local climate and weather variability~\cite{Apa12} and the structure of customers. 
Factors that disturb electricity demand time series include unpredictable economic events, extreme weather conditions and political decisions~\cite{Dog16}.

\begin{figure}[t]
\centering
\includegraphics[width=0.8\textwidth]{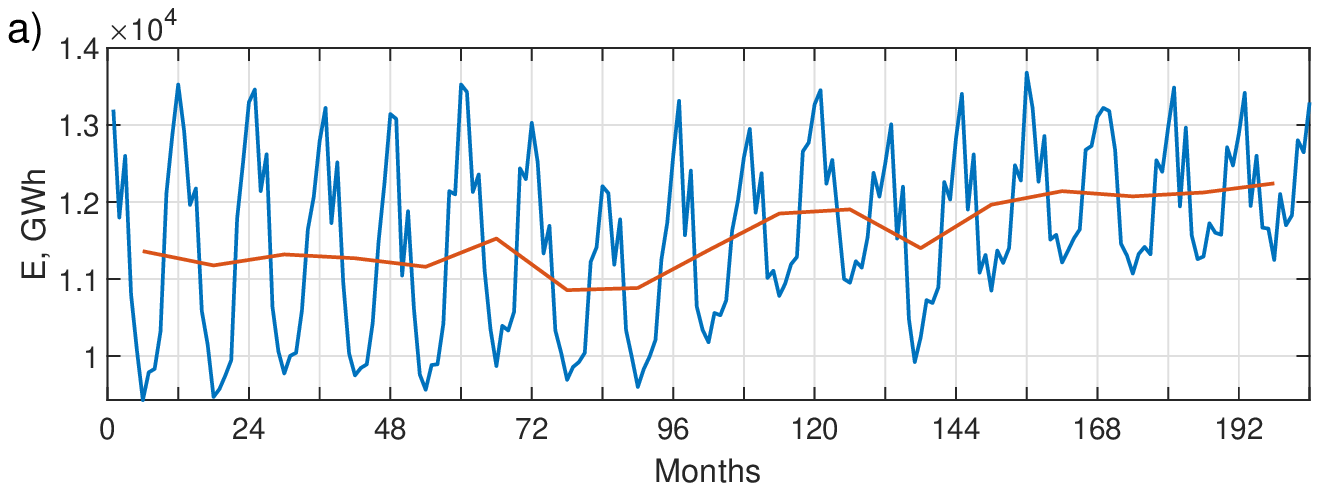}
\includegraphics[width=0.365\textwidth]{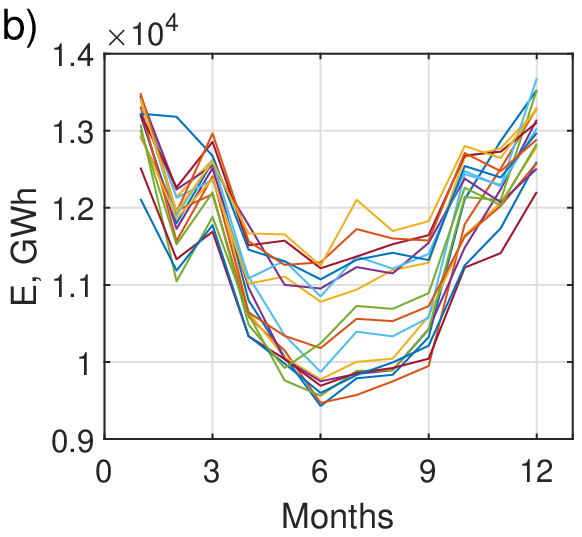}
\includegraphics[width=0.435\textwidth]{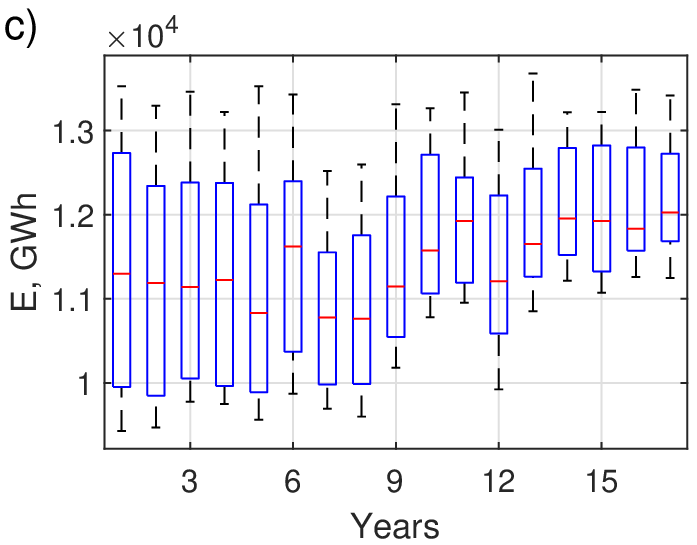}
\caption{Monthly electricity demand time series for Poland (a), its yearly patterns (b), and dispersion in successive years (c).}
\label{fig:PL_TS}
\end{figure}

\label{PS}

\begin{figure}[t]
\centering
\includegraphics[width=0.8\textwidth]{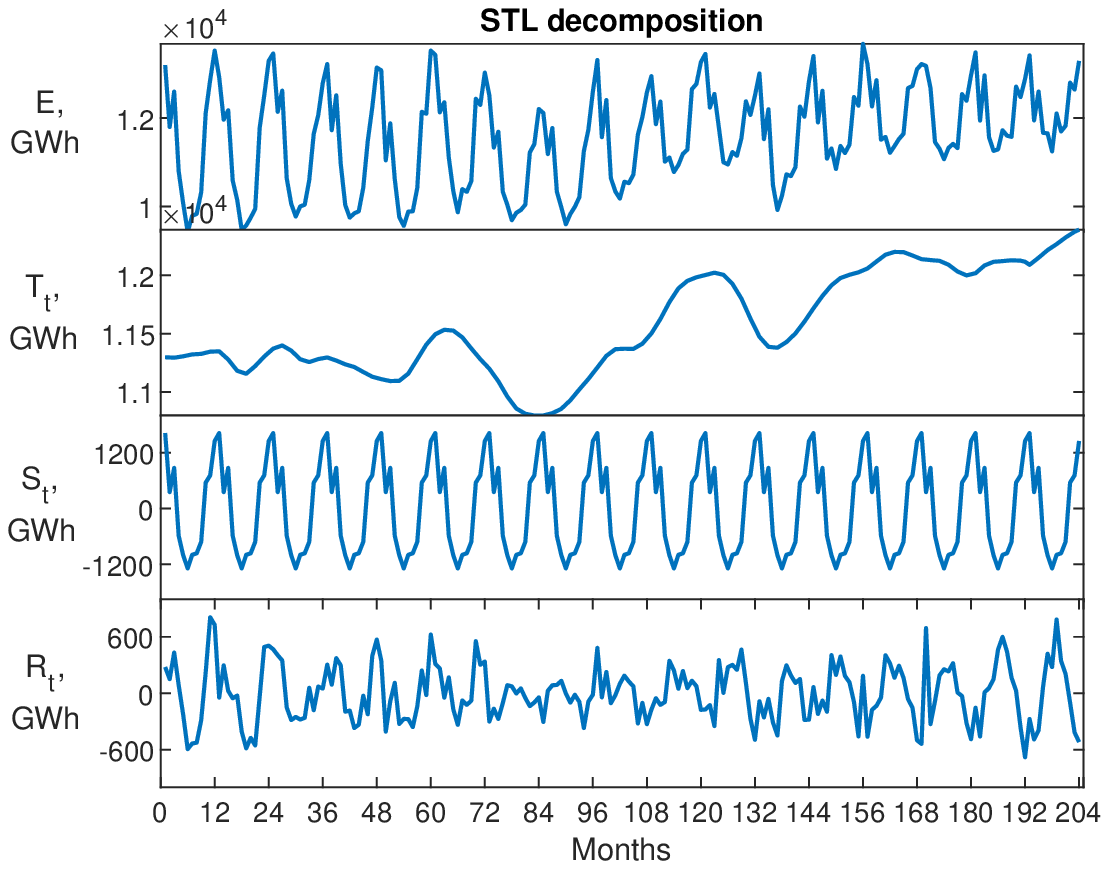}
\caption{STL decomposition of the monthly electricity demand time series for Poland.}
\label{fig:PL_STL}
\end{figure}

Figure~\ref{fig:PL_TS} provides an example of the monthly electricity demand time series. From this figure we can observe an upward trend and changing yearly patterns over time. 
Additionally, the dispersion of the yearly cycles changes significantly over time, from $\sigma = 696$ to $1484$ MWh. Decomposition of this time series using the STL method (seasonal and trend decomposition using Loess~\cite{Cle90}; see Fig. \ref{fig:PL_STL}) reveals a strong seasonal component ($S_t$). The ratio of its variance to the total variance of the series is 77\%. This ratio is 16\% for the trend ($T_t$) and 7\% for the random component ($R_t$).

\begin{figure}[t]
\centering
\includegraphics[trim=3cm 2cm 3cm 1cm, clip=true, width=1\textwidth]{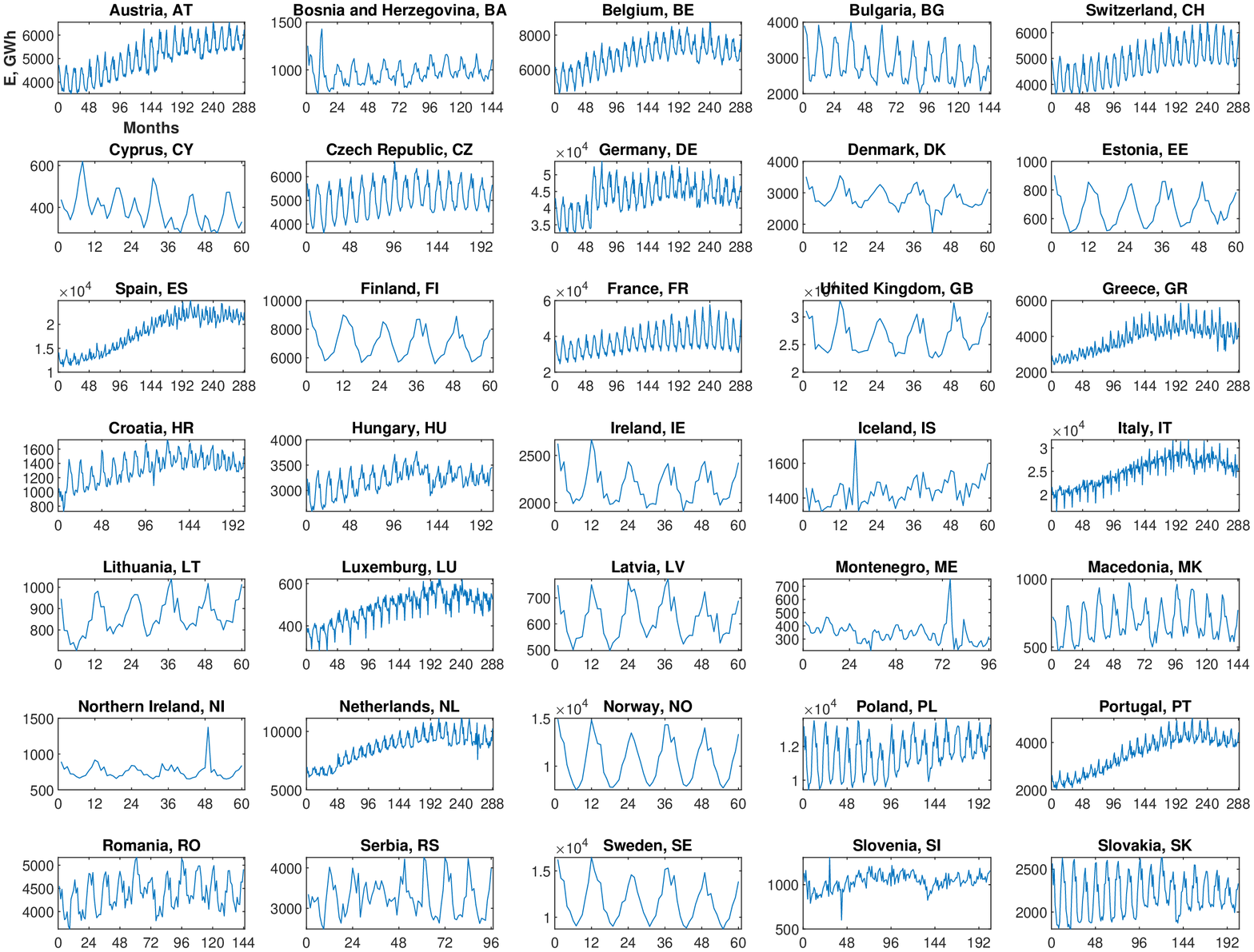}
\caption{Monthly electricity demand time series for European countries. Note that the y-axis unit is GWh for all plots.}
\label{fig:TS}
\end{figure}

All the time series included in the dataset are presented in Fig. \ref{fig:TS}. They differ substantially in: 
\begin{itemize}
    \item level -- mean monthly demand varies from 343 (ME) to 43702 MWh (DE), 
    \item dispersion -- mean yearly standard deviation varies from 72 (LU) to 6581 MWh (FR),
    \item autocorrelation -- lag 12 autocorrelation (yearly period) varies from 0.09 (ME) to 0.92 (CH),
    \item share of the trend, seasonal and random components -- the countries with the highest share of the trend (over 80\%) are ES, PT, NL and IT, those with the highest share of the seasonal component (over 90\%) are NO, FI, EE, SE and IE, and those with a high share of the random component (over 30\%) are ME, NI and RS, 
    \item length -- from 5 (12 countries) to 24 years (11 countries), and 
    \item similarity of the yearly pattern.
\end{itemize}

Constructing the forecasting model for such time series is a challenging task. This problem becomes especially difficult when the time series is short and contains strong random fluctuations and irregular spikes, as is the case for the BA, DK, IS, ME, NI and SI time series (see Fig. \ref{fig:TS}).

\subsection{Training and Evaluation Setup}

The dataset is split into train, validation and test subsets. The test subset is constructed by cutting the last horizon (twelve months of 2014) off each of the 35 time series. The validation and train subsets are obtained by splitting the full train sets at the boundary of the last horizon of each time series. Thus, we treat the twelve months of 2013 as a validation subset. We use the train and validation subsets to tune hyperparameters. Once the hyperparameters are determined, we train the model on the full train set and report results on the test set.

\begin{table}[t]
    \centering
    \caption{Settings of N-BEATS hyperparameters and the hyperparameter search grid.}
    \label{table:hyperparameter_settings}

    \begin{tabular}{lcc}
        \toprule
        Hyperparameter & Value  & Grid \\ 
        \midrule
        Epochs	& 20 & 20	\\ 
        Batches per epoch	& 50 & [25, 50, 100]	\\ 
        Loss & $\pmape$ & $\pmape$ \\
        $\tau$ & 0.35 & [0.3, 0.35, 0.4, 0.45, 0.5, 0.55, 0.6] \\
        Width ($d_h$) & 512 & [256, 512, 1024]	  \\
        Blocks ($R$) & 3 & [1, 2, 3, 5, 10]	  \\
        Layers ($L$) & 3 & [2, 3, 4]  \\
        Sharing & True & [True, False]	  \\
        Lookback period ($w$, months) & 12 & [6, 9, 12, 24]  \\
        Batch size & 256 & [128, 256, 512, 1024]	\\
        Optimizer & Adam & Adam \\
        Learning rate & 0.001 & 0.001 \\
        \bottomrule
    \end{tabular}
\end{table}

The N-BEATS model described in detail in Section~\ref{N-BEATS} is evaluated in this section using the hyperparameter settings presented in Table~\ref{table:hyperparameter_settings}. This table also provides the hyperparameter ranges that were used during hyperparameter tuning. All the hyperparameters were adjusted by minimizing $\mape$ on the validation set, with the exception of the $\pmape$ loss parameter $\tau$, which was selected on the validation set to minimize forecasting bias. We do not use weight decay; instead, regularization is achieved via an ensemble of 64 models. Each of the models in the ensemble is trained using a different random initialization and a different random sequence of batches. The objective function used to train the network is pinball $\mape$ with $\tau=0.35$, which is described in Section~\ref{ssec:pinball_mape} (see eq.~\eqref{eqn:pinball_mape}), averaged over all forecasts in the batch within horizon $H=12$.  

The model is trained using the Adam optimizer with default TensorFlow 2.0 settings and an initial learning rate of 0.001 for 20 epochs. The learning rate is annealed by a factor of 2 every 2 epochs starting at epoch 15. One epoch consists of 50 batches of size 256, and the model takes the history of 12 points (12 months; $w=12$) and predicts 12 points (12 months; $H=12$) ahead in one shot. Each training batch is assembled using weighted stratified sampling over time series IDs. First, 256 time series IDs are sampled with replacement, and the probability of sampling a given time series is proportional to the length of the time series. Second, the split time point is chosen uniformly at random for each of the time series IDs sampled in the previous step. 

The weighted stratified sampling of time series is important because each time series has a different length. A training sample is formed by splitting a given time series at a split point, feeding the history window preceding the split point into the network and computing a loss using the values following the split point. Obviously, a smaller time series will generate a smaller number of unique training samples. Therefore, we should not sample time series IDs uniformly. If we do, then each sample from a short time series will be used to adjust the training loss more times on average and the model will be overfitting more on the shorter time series than on the longer ones. Weighted stratified sampling solves this problem by making sure that each training sample is used to adjust training loss the same number of times on average\footnote{Similar effect can be achieved by creating all viable training samples from all time series and putting them in a flat table. The batches can then be assembled by uniformly sampling the rows of the flat table. For a simple in-memory data loader, this is appropriate for smaller datasets but may quickly inflate RAM usage for larger datasets.}.

Due to the stochastic nature of N-BEATS, all results reported for this model take averages over 100 trials. In each trial, we build an ensemble of 64 models bootstrapped from the set of 1024 trained models.

\subsection{Baseline Models}

The baseline models that we use in our comparative studies are outlined below. The hyperparameters of the baselines were selected on the training set in grid search procedures. For more details about the models and their hyperparameter settings, please refer to \ref{sec:baseline_models_hyperparameters}.

\begin{itemize}

    \item ARIMA and ETS -- classical statistical models \texttt{auto.arima} and \texttt{ets} from R package \texttt{forecast}. Both models use the Akaike information criterion (AICc)~\cite{Hyn20} for model structure and order selection.

	\item $k$-NNw+ETS, FNM+ETS, N-WE+ETS, GRNN+ETS -- hybrid models that combine either $k$-nearest neighbor weighted regression, fuzzy neighborhood model, Nadaraya–Watson estimator or general regression NN for yearly cycle forecasting and ETS for mean yearly load and dispersion forecasting~\cite{Dud20a}. 
	
	\item MLP -- perceptron with a single hidden layer and sigmoid non-linearities~\cite{Pel19b}. 
	
	\item ANFIS -- a standard adaptive neuro-fuzzy inference system~\cite{Pel18}.

	\item LSTM -- a standard LSTM model~\cite{Pel20}.
	
	\item ETS+RD-LSTM -- a hybrid residual-dilated LSTM and ETS model~\cite{Dud21}.
	This model combines ETS, advanced LSTM and ensembling.

\end{itemize}

\subsection{Results}

Forecasting quality metrics averaged over the 35 countries are presented in Table~\ref{tab1}. They include: median of absolute percentage error ($\ape$), $\mape$, interquartile range of $\ape$ ($\iqr$) as a measure of forecast dispersion, root mean square error ($\rmse$) and mean percentage error ($\mpe$). 
Note the lowest values for each error measure and $\iqr$ for N-BEATS. It clearly outperforms all the other models in terms of accuracy. N-BEATS having a $\mape$ below 4\% should be considered a major achievement. The second most accurate model is N-WE+ETS with a $\mape = 4.37\%$. Other hybrid models combining ETS and machine learning had similar errors, below 4.5\%.
The best results for N-BEATS were confirmed by computing bootstrapped confidence intervals for the difference in the $\mape$ metric between the baseline methods and N-BEATS. None of the 99\% confidence intervals overlap zero (see the $\mape$ $\textsc{diff}$ column in Table~\ref{tab1}). Therefore, we conclude that the difference in $\mape$ between N-BEATS and the other models is statistically significant at level $\alpha=0.01$.

$\mpe$, shown in Table~\ref{tab1}, allows us to assess the bias of the forecasts produced by the proposed and baseline models. All the models produced negatively biased forecasts, which means overprediction. Note that N-BEATS has the lowest bias, with an $\mpe=-0.34\%$, while the biases for the other models exceed --1\%. In the case of N-BEATS, the $t$-test did not reject the null hypothesis that $\pe$ comes from a normal distribution with zero mean ($\alpha=0.01$). All the other models failed this test. Therefore, it can be concluded that N-BEATS is the only model that produced unbiased forecasts. Note that N-BEATS has a mechanism to deal with bias included in the loss function $\pmape$ \eqref{eqn:pinball_mape}. $\pmape$ asymmetry is controlled by parameter $\tau$. Its optimal value was selected as 0.35, which allowed the model to reduce the negative bias significantly. 

\begin{table}[]
	\caption{Forecasting metrics. All $\mape$ difference results between N-BEATS and the other algorithms are statistically significant at the 1\% level. This follows from the $\mape$~$\textsc{diff}$ column, which shows the mean difference in $\ape$ between the baseline algorithms and N-BEATS, accompanied by the 99\% confidence intervals. The upper confidence interval boundary is shown in superscript and the lower boundary in subscript in the $\mape$~$\textsc{diff}$ column. The confidence intervals are computed using a 100k-sample bootstrap sampled with replacement from the difference in $\ape$ between the baseline algorithms and N-BEATS.}
	\label{tab1}
	\setlength{\tabcolsep}{6.4pt}
	\centering
	\begin{tabular}{lcccccc}
		\toprule
		Model       & \multicolumn{1}{c}{$\textsc{median}$ $\ape$} & \multicolumn{1}{c}{$\mape$} & \multicolumn{1}{c}{$\iqr$} & \multicolumn{1}{c}{$\rmse$} & \multicolumn{1}{c}{$\mpe$} & \multicolumn{1}{c}{$\mape$ $\textsc{diff}$} \\ \midrule \vspace{0.1cm}
		
		ARIMA	&	3.32	&	5.65	&	5.24	&	463.07 & --2.35 & $1.87 \ ^{3.10}_{1.01}$ \\ 
		\vspace{0.1cm}
		ETS	&	3.50	&	5.05	&	4.80	&	374.52 & --1.04	& $1.27 \ ^{1.76}_{0.81}$ \\
		\vspace{0.1cm}
		k-NNw+ETS	&	2.71	&	4.47	&	3.52	&	327.94 & --1.25	& $0.69 \ ^{1.25}_{0.18}$  \\ \vspace{0.1cm}
		FNM+ETS	&	2.64	&	4.40	&	3.46	&	321.98 & --1.26	& $0.63 \ ^{1.19}_{0.14}$ \\ \vspace{0.1cm}
		N-WE+ETS	&	2.68	&	4.37	&	3.36	&	320.51 & --1.26	& $0.59 \ ^{1.14}_{0.11}$ \\ \vspace{0.1cm}
		GRNN+ETS	&	2.64	&	4.38	&	3.51	&	324.91 & --1.26	& $0.61 \ ^{1.14}_{0.13}$ \\ \vspace{0.1cm}
		MLP	&	2.97	&	5.27	&	3.84	&	378.81  & --1.37 & $1.49 \ ^{2.62}_{0.72}$	\\ \vspace{0.1cm}
		ANFIS	&	3.56	&	6.18	&	4.87	&	488.75 & --2.51	& $2.40 \ ^{3.56}_{1.41}$ \\ \vspace{0.1cm}
		LSTM	&	3.73	&	6.11	&	4.50	&	431.83 & --3.12	& $2.33 \ ^{2.98}_{1.73}$ \\ \vspace{0.1cm}
		ETS+RD-LSTM	&	2.74	&	4.48	&	3.55	&	347.24 & --1.11	& $0.70 \ ^{1.40}_{0.15}$ \\  \vspace{0.1cm}	 
		N-BEATS	&	\textbf{2.55}	&	\textbf{3.78}	&	\textbf{3.30}	&	\textbf{309.91} & \textbf{--0.34}	& -- \\ 	
		\bottomrule
	\end{tabular}
\end{table}

The consistency of our model is evaluated in Table~\ref{tab2}. To verify the consistency, we trained N-BEATS 1024 times with different random seed values. From the sample of 1024 trained models we then selected 100 ensembles containing 64 models each via bootstrapping. More specifically, each of the 100 ensembles was created by sampling 64 of the 1024 models with no replacement. The $\mape$ and $\mpe$ distributions over 100 trials appear in Table~\ref{tab2}. Additionally, in Table~\ref{tab2} we compare the error distributions achieved by N-BEATS with two training losses: pinball loss and the proposed $\pmape$. Note the much lower error values for the proposed model with $\pmape$: the highest $\mape$ and $\mpe$ values achieved for $\pmape$ are lower then the corresponding minimal errors achieved for N-BEATS with pinball loss. 
N-BEATS is quite stable in both versions, i.e., the mean errors obtained in the individual runs are narrowly distributed (see the low Std values and tight confidence intervals in Table~\ref{tab2}). This is because N-BEATS is an ensemble-based model combining 64 base models. A great advantage of ensemble learning is reducing the variance of predictions as well as generalization error. If we compare a standard deviation of errors for the pool of individual base models and for the ensemble model we observe a significant reduction: 

\begin{itemize}
\item for N-BEATS with $\pmape$, Std $\mape$ was reduced from 0.2102 to 0.0303, and Std $\mpe$ from 0.3873 to 0.0479, and
\item for N-BEATS with pinball loss, Std $\mape$ was reduced from 0.1655 to 0.0216, and Std $\mpe$ from 0.3454 to 0.0392.
\end{itemize}

\begin{table}[]
	\caption{$\mape$ and $\mpe$ distributions for N-BEATS.}
	\label{tab2}
	\centering
\begin{tabular}{lcc|lcc}
\toprule
\multicolumn{1}{c}{$\mape$} & \multicolumn{1}{r}{N-BEATS} & N-BEATS      & \multicolumn{1}{c}{$\mpe$} & N-BEATS & N-BEATS      \\
\multicolumn{1}{c}{\textbf{}}    & \multicolumn{1}{r}{$\pmape$}  & pinball loss & \multicolumn{1}{c}{\textbf{}}   & $\pmape$  & pinball loss \\ \midrule
Mean                             & 3.78                        & 4.01         & Mean                            & --0.34    & --0.80         \\
Std                              & 0.0303                      & 0.0216       & Std                             & 0.0479  & 0.0392       \\
Min                              & 3.70                        & 3.96         & Min                             & --0.23    & --0.73         \\
5\%                              & 3.74                        & 3.97         & 5\%                             & --0.25    & --0.75         \\
25\%                             & 3.75                        & 3.99         & 25\%                            & --0.31    & --0.78         \\
50\%                             & 3.78                        & 4.01         & 50\%                            & --0.34    & --0.80         \\
75\%                             & 3.80                        & 4.02         & 75\%                            & --0.37    & --0.83         \\
95\%                             & 3.83                        & 4.04         & 95\%                            & --0.41    & --0.87         \\
Max                              & 3.85                        & 4.06         & Max                             & --0.47    & --0.91         \\ 
\bottomrule
\end{tabular}
\end{table}

\begin{figure}[t]
\centering
\includegraphics[width=1\textwidth]{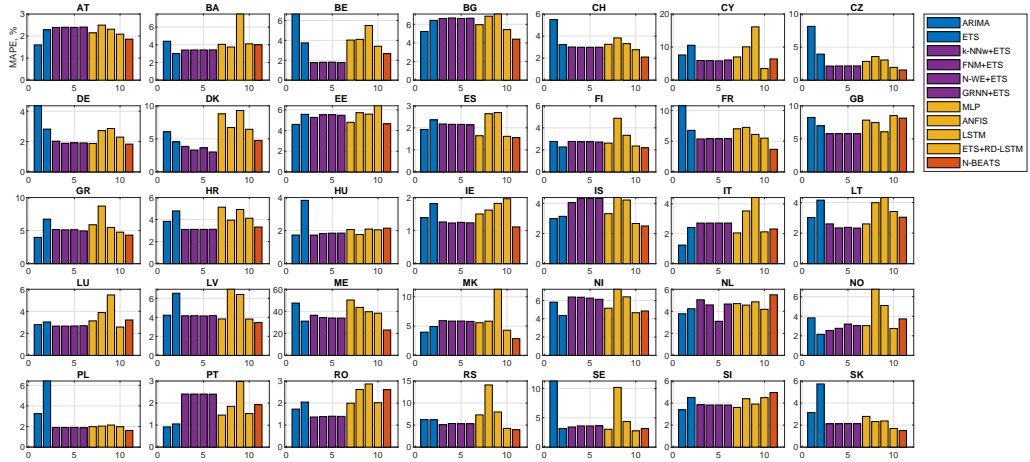}
\caption{$\mape$ for each country. Note that the y-axis unit is percent for all plots.}
\label{fig:MAPE_countries}
\end{figure}

\begin{figure}[t]
\centering
\includegraphics[width=0.7\textwidth]{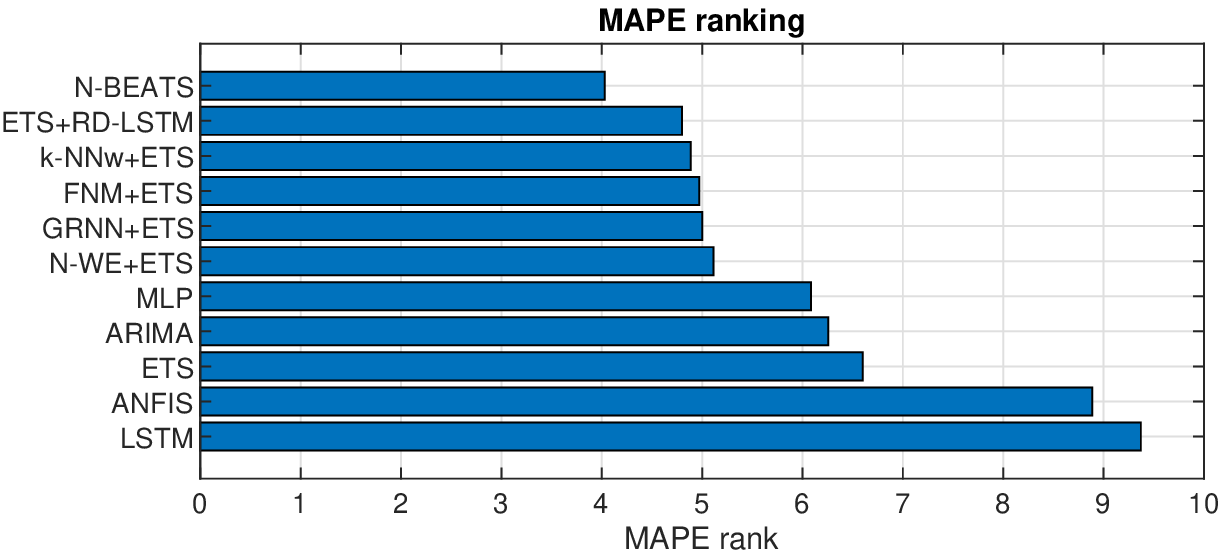}
\includegraphics[width=0.7\textwidth]{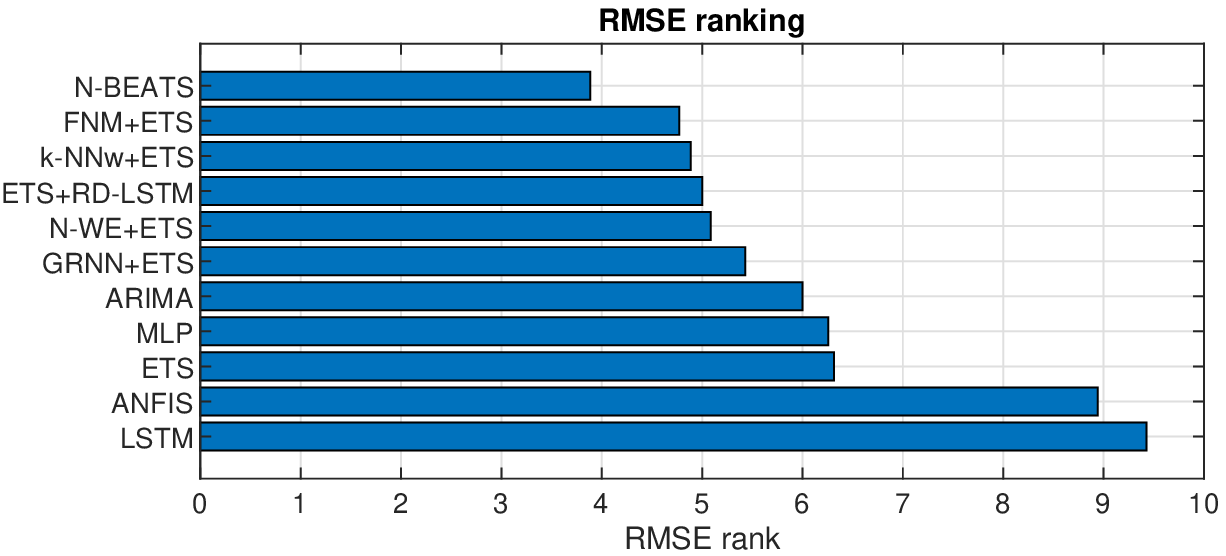}
\caption{Rankings of the MTLF models.}
\label{fig:MAPE_ranking}
\end{figure}

Figure~\ref{fig:MAPE_countries} shows $\mape$ error for each country. As we can see from this figure, N-BEATS is one of the most accurate models for most countries. In fact, it outperforms all the other models in 16 out of the 35 cases. The models' average ranks from the rankings for individual countries are shown in Fig.~\ref{fig:MAPE_ranking}. The rankings were performed for $\mape$ and $\rmse$. In both cases, N-BEATS is in first position with a large advantage over the other models. 

Fig. \ref{fig:MAPE_ranking} shows mean error for each month of the test period. Characteristically, for this dataset, electricity demand in August-October is predicted with lower error, and demand in February is predicted with the highest error. Note the excellent results of N-BEATS, which produces the most accurate forecasts for 7 months. For February, it gives a $\mape=4.74\%$, while the second-best model, ARIMA, gives a $\mape=5.88\%$. 

\begin{figure}[t]
\centering
\includegraphics[width=0.7\textwidth]{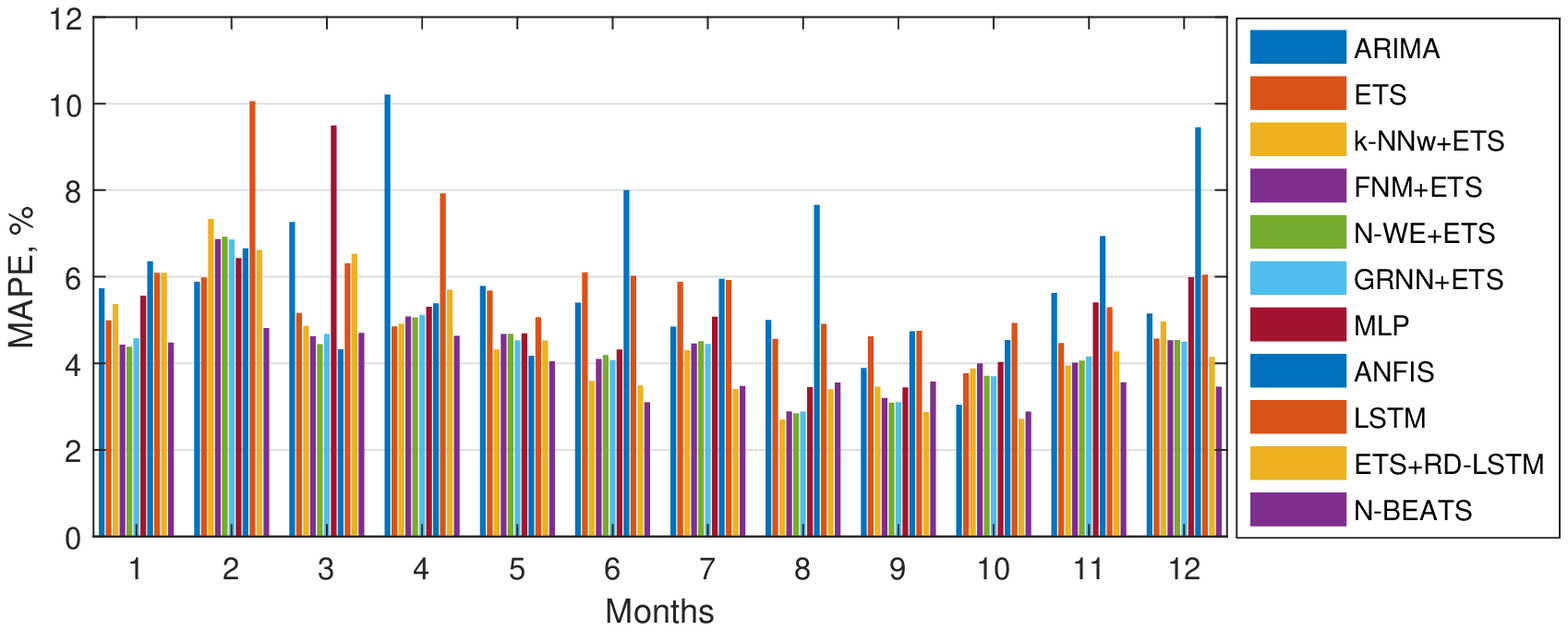}
\caption{$\mape$ for each month of the test period.}
\label{fig:MAPE_month}
\end{figure}

\begin{figure}[t]
\centering
\includegraphics[width=0.31\textwidth]{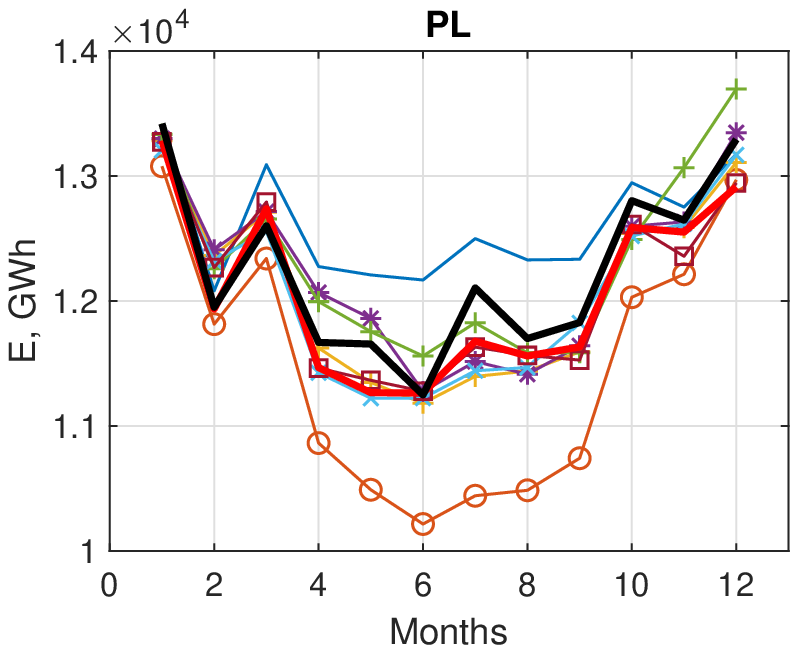}
\includegraphics[width=0.31\textwidth]{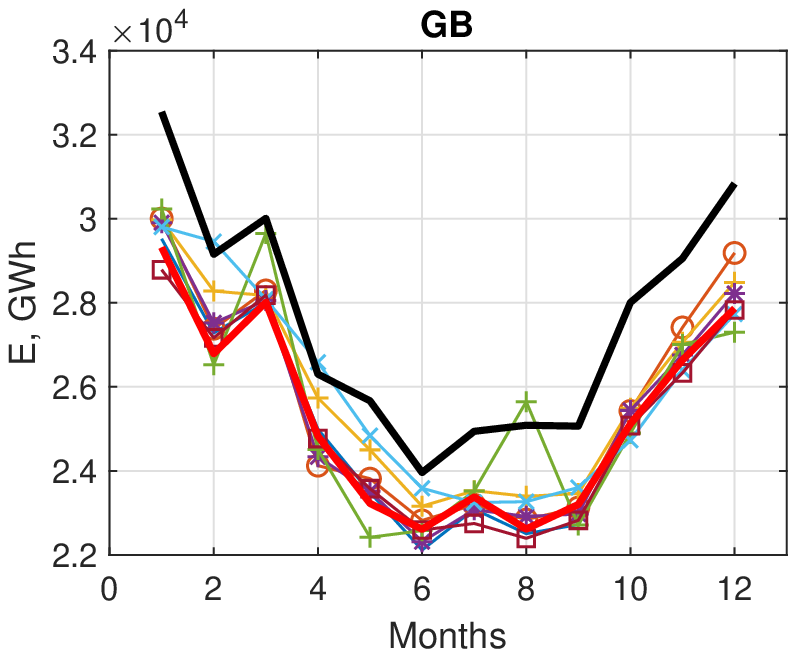}
\includegraphics[width=0.31\textwidth]{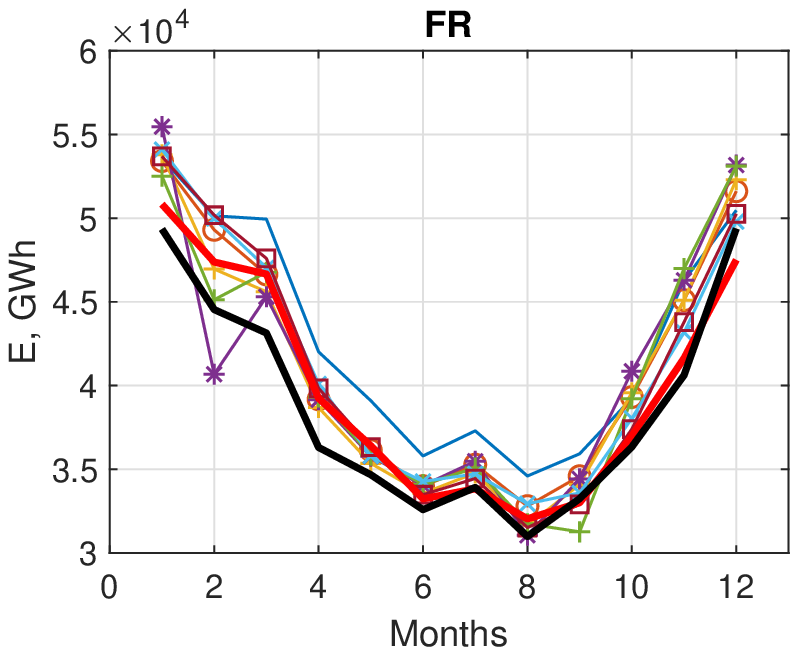}
\includegraphics[width=0.31\textwidth]{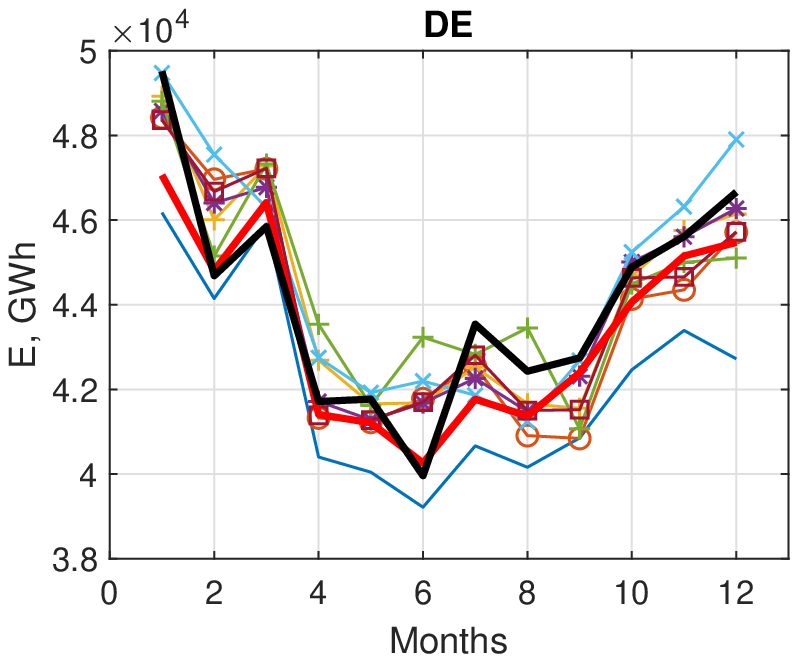}
\includegraphics[width=0.31\textwidth]{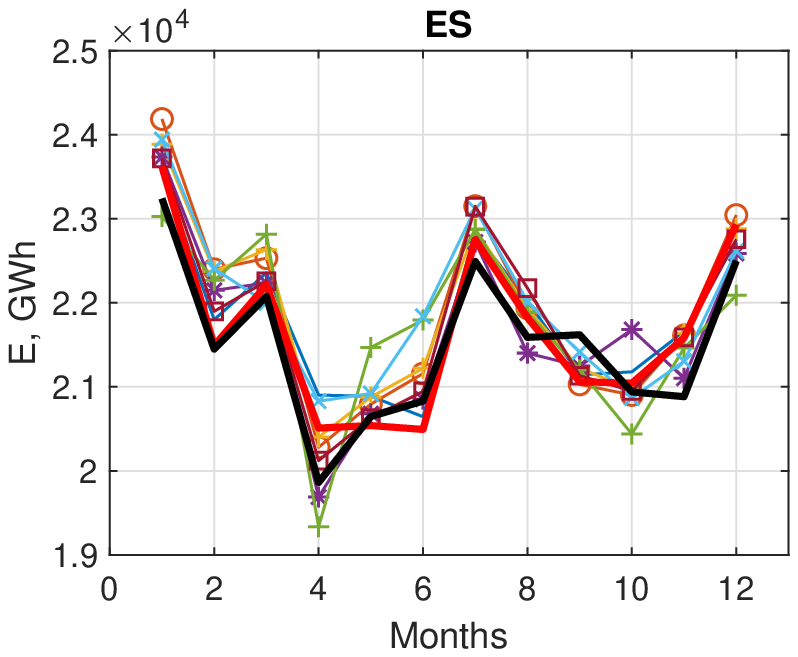}
\includegraphics[width=0.31\textwidth]{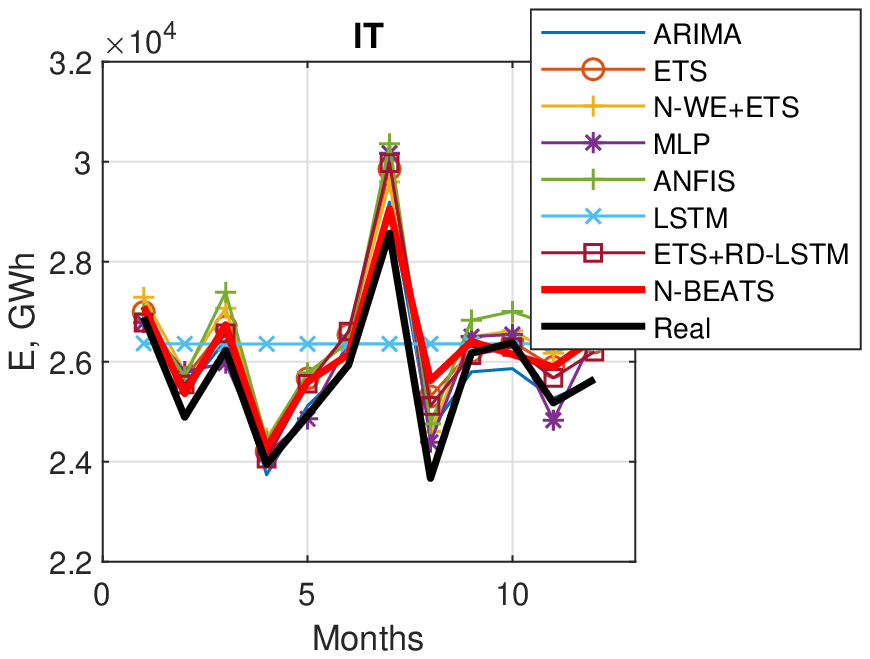}
\caption{Examples of forecasts.}
\label{fig:Forecasts}
\end{figure}

Examples of forecasts for selected countries are depicted in Fig. \ref{fig:Forecasts}. For PL, FR, DE and ES, N-BEATS produced the most accurate forecasts. Note the outlier forecasts of LSTM for IT and the classical models, ARIMA and ETS, for PL. For GB, the forecasts of all models were underestimated. This results from the fact that demand went up unexpectedly in 2014 despite the downward trend observed from 2010 to 2013. The reverse situation for FR caused a slight overestimation of forecasts. For GB, N-BEATS was one of the least accurate models with a $\mape = 8.10\%$.  

\subsection{Discussion}

The results presented in Subsection 3.4 clearly show N-BEATS has the best performance over statistical, classical machine learning and hybrid methods. It outperforms the baseline models in terms of accuracy and unbiased forecast distribution. 
The success of N-BEATS is attributable to backward and forward residual links, a deep stack of fully connected layers and ensembling. The architecture can be applied without modification to a wide range of target domains, including MTLF, which was confirmed in this study.

N-BEATS does not require decomposition of the time series or any data preprocessing. Many statistical and machine learning approaches do not work with time series exhibiting non-stationarity, non-linear relationships between input and output variables, or seasonal variations. They require additional preliminary steps such as differencing, detrending, deseasonalization or decomposition. Sometimes these procedures are included in the model structure, as in the case of ETS or similarity-based methods \cite{Dud20a}. N-BEATS deals with raw time series and processes them properly using built-in mechanisms such as non-linear mapping on several levels, residual links, forecast and backcast paths, and aggregation of the partial forecasts. This, together with the final ensembling, leads to accurate forecasts.

In this work, we modify the original N-BEATS implementation by introducing the pinball-$\mape$ loss function \eqref{eqn:pinball_mape}. It allows N-BEATS to directly minimize $\mape$, which we selected as the main MTLF performance metric, and to reduce forecast bias. When compared to the standard pinball loss function, $\pmape$ significantly reduces both $\mape$ and forecast bias (see Table~\ref{tab2}). Note that N-BEATS can implement any loss function ($\smape$, $\rmse$, etc.) in a pinball version. This allows the model to be optimized for any forecasting problem with a specific quality metric incorporating the bias.

In this study, we confirmed that training a deep learning model on multiple time series (cross-learning) successfully leads to transferring and sharing individual learnings. All the other models except ETS+RD-LSTM are trained and optimized separately for a single time series. Cross-learning enables the method to capture the shared features and components of the time series. It also speeds up learning and optimization of the model, which is especially important for complex deep learning models with a huge number of parameters and hyperparameters.

N-BEATS was proposed in two configurations: generic and interpretable \cite{Ore19}. In this study, we use a generic variant with the aim of validating the hypothesis that the generic deep learning approach performs exceptionally well on the MTLF problem using no domain knowledge. The interpretable N-BEATS configuration forces a deep learning model to decompose its forecast into distinct human-interpretable outputs, i.e., trend and seasonal components. Future work will explore the usefulness of MTLF using interpretable N-BEATS for power system operators and practitioners. 


\section{Conclusions} \label{Con}

Accurate load forecasts are of great importance in ensuring safe and efficient power system operation, increasing electricity market revenues and reducing financial risks. The mid-term load forecasting considered in this work is a challenging problem that requires the forecasting model to be highly flexible and to deal with non-stationarity and seasonality. In this study, we proposed and empirically validated a new architecture for mid-term electricity load forecasting problem that responds to these expectations -- the N-BEATS neural network. 

The empirical study of the mid-term electricity load forecasting  for 35 European countries showed N-BEATS had the best performance over statistical, machine learning and hybrid methods. N-BEATS clearly outperformed its competitors in terms of both accuracy and forecast bias. As for mean absolute percentage error, our method provided a relative gain of 25\% with respect to the statistical methods, 28\% with respect to machine learning methods and 14\% with respect to the advanced state-of-the-art hybrid methods. Its success is due to a unique architecture that combines a deep stack of fully connected layers, backward and forward residual links, aggregation of the partial forecasts in a hierarchical fashion and ensembling. Cross-learning, i.e., learning on multiple time series, enables N-BEATS to capture the shared features and components of the individual time series. A great advantage of N-BEATS is in being able to deal with the raw time series, without requiring their decomposition or any preprocessing. 

In our implementation of N-BEATS for mid-term electricity load forecasting problem, we introduced the pinball mean absolute percentage error loss function, which allows the model to directly minimize the main mid-term electricity load forecasting problem performance metric and to reduce forecast bias. It is worth noting that N-BEATS can implement any loss function in a pinball version. Therefore, the model can be optimized for any forecasting problem with a specific quality metric incorporating the bias.











\bibliographystyle{elsarticle-num-names}
\bibliography{main}

\clearpage
\appendix

\section{N-BEATS TensorFlow implementation} \label{sec:nbeats-implementation}

\begin{listing}[h]
\begin{minted}{python}
import tensorflow as tf


class NBEATSBlock(tf.keras.layers.Layer):
    def __init__(self, input_size: int, output_size: int, block_layers: int, hidden_units: int):
        super().__init__()
        self.fc_layers = []
        for i in range(block_layers):
            self.fc_layers.append(tf.keras.layers.Dense(hidden_units,
                                                        activation=tf.nn.relu))
        self.forecast = tf.keras.layers.Dense(output_size, activation=None)
        self.backcast = tf.keras.layers.Dense(input_size, activation=None)

    def call(self, x):
        inputs = x
        for layer in self.fc_layers:
            x = layer(x)
        backcast = tf.keras.activations.relu(inputs - self.backcast(x))
        return backcast, self.forecast(x)


class NBEATS(tf.keras.layers.Layer):
    def __init__(self, input_size: int, output_size: int, block_layers: int, hidden_units: int,
                 num_blocks: int, block_sharing: bool):
        super().__init__()
        self.blocks = [NBEATSBlock(input_size=input_size, output_size=output_size,
                                   block_layers=block_layers, hidden_units=hidden_units)]
        for i in range(1, num_blocks):
            if block_sharing:
                self.blocks.append(self.blocks[0])
            else:
                self.blocks.append(NBEATSBlock(input_size=input_size, output_size=output_size,
                                               block_layers=block_layers, hidden_units=hidden_units))

    def call(self, x):
        level = tf.reduce_max(x, axis=-1, keepdims=True)
        backcast = tf.math.divide_no_nan(x, level)
        forecast = 0.0
        for block in self.blocks:
            backcast, forecast_block = block(backcast)
            forecast = forecast + forecast_block
        return forecast * level


nbeats = NBEATS(input_size=12, output_size=12, block_layers=3, num_blocks=3, hidden_units=512, block_sharing=True)
inputs = tf.random.normal([256, 12])
forecast = nbeats(inputs)
\end{minted}
\caption{N-BEATS TensorFlow inference code.}
\label{listing:nbeats-inference-code}
\end{listing}

\clearpage
\section{Hyperparameters of the baseline models} \label{sec:baseline_models_hyperparameters}

The detailed descriptions of the baseline models are provided below, and the hyperparameter settings appear in Tables~\ref{tabA1}--\ref{tabA4}.

\begin{itemize}

	\item ARIMA -- ARIMA$(p, d, q)(P, D, Q)_{12}$ model implemented in the function \texttt{auto.arima} in R environment (package \texttt{forecast}). This function implements automatic modeling to obtain the optimal ARIMA model with regard to AICc. 

	\item ETS -- exponential smoothing state space model \cite{Hyn08} implemented in the function \texttt{ets} (R package \texttt{forecast}). This implementation includes many types of ETS models depending on how the seasonal, trend and error components are taken into account. As in the case of \texttt{auto.arima}, \texttt{ets} returns the optimal model, estimating its parameters using AICc.

	\item $k$-NNw+ETS -- a hybrid model combining $k$-nearest neighbor weighted regression and ETS \cite{Dud20a}. It uses the pattern representation of time series. Patterns, which express unified yearly cycles, are forecasted using $k$-nearest neighbor with a linear weighted function. Mean yearly load and yearly dispersion are both forecasted using ETS. The model hyperparameters are: input pattern length and number of nearest neighbors $k$. 
	
	\item FNM+ETS -- a hybrid model combining the fuzzy neighborhood model for pattern forecasting and ETS for yearly mean and dispersion forecasting \cite{Dud20a}. The model hyperparameters are: input pattern length and membership function width.
	
	\item N-WE+ETS -- a hybrid model combining the Nadaraya–Watson estimator for pattern forecasting and ETS for yearly mean and dispersion forecasting \cite{Dud20a}. The model hyperparameters are: input pattern length and kernel bandwidth parameters.
	
	\item GRNN+ETS -- a hybrid model combining the general regression NN for pattern forecasting and ETS for yearly mean and dispersion forecasting \cite{Dud20a}. The model hyperparameters are: input pattern length and node bandwidth parameters.
	
	\item MLP -- multilayer perceptron with a single hidden layer with sigmoid neurons \cite{Pel19b}. It works on pattern representation of the time series and uses the Levenberg-Marquardt learning method with Bayesian regularization to prevent overfitting. The MLP hyperparameters are: input pattern length and number of hidden nodes.  
	
	\item ANFIS -- adaptive neuro-fuzzy inference system \cite{Pel18}. A hybrid learning method is applied for ANFIS training that combines the least-squares and backpropagation gradient descent methods. The ANFIS hyperparameters are: input pattern length and number of rules.

	\item LSTM -- long short-term memory \cite{Pel20}. A standard LSTM model is used without time series preprocessing. LSTM was optimized using the Adam optimizer. The length of the hidden state was the only hyperparameter to be tuned. The other hyperparameters remain at their default values.
	
	\item ETS+RD-LSTM -- a hybrid residual-dilated LSTM and ETS model \cite{Dud21}.
	This model combines ETS, advanced LSTM and ensembling. The model hyperparameters are: number of epochs = 16, learning rate = 0.001, length of the cell and hidden states = 40, asymmetry parameter in pinball loss = 0.4, regularization parameter = 50 and ensembling parameters $L=5, K=3, R=3$ (see \cite{Dud21} for details). 

\end{itemize}

\begin{table}[]
\small
  \centering
  \setlength{\tabcolsep}{3.1pt}
  \caption{Hyperparameters for the baseline models: ARIMA, ETS, k-NNw+ETS, FNM+ETS, N-WE+ETS, GRNN+ETS and LSTM. Here, TS is the time series ID. For ARIMA, $p, d, q$ are hyperparameters of the non-seasonal part: order of the auto-regressive part, degree of differentiator and order of the moving average, respectively, and $P, D, Q$ are hyperparameters of the seasonal part: order of the auto-regressive part, degree of differentiator and order of the moving average, respectively. For ETS, error, trend and seas. (seasonal) are components of ETS, and M, N, A and Ad - multiplicative, none, additive and additive damped, respectively - denote the way each component is integrated in the model. For the other algorithms, $n$ is input pattern length, $k$ is the number of nearest neighbors, and $a, b$ are bandwidth parameters.}
   \label{tabA1}%
    \begin{tabular}{c|cccccc|ccc|cc|cc|cc|cc|c}
   
    \toprule
    \multicolumn{1}{c|}{TS} & \multicolumn{6}{c|}{ARIMA}                    & \multicolumn{3}{c|}{ETS} & \multicolumn{2}{c|}{k-NNw} & \multicolumn{2}{c|}{FNM} & \multicolumn{2}{c|}{N-WE} & \multicolumn{2}{c|}{GRNN} & LSTM \\

    \multicolumn{1}{c|}{} & \multicolumn{6}{c|}{}                    & \multicolumn{3}{c|}{} & \multicolumn{2}{c|}{+ETS} & \multicolumn{2}{c|}{+ETS} & \multicolumn{2}{c|}{+ETS} & \multicolumn{2}{c|}{+ETS} &  \\
    
          & $p$     & $d$     & $q$     & $P$     & $D$     & $Q$     & error & trend & seas. & $n$     & $k$     & $n$     & $a$     & $n$     & $b$     & $n$     & $a$     & \#nodes  \\ 
\midrule
   
    AT    & 3     & 1     & 3     & 1     & 0     & 0     & M     & N     & A     & 24    & 19    & 24    & 0.18  & 24    & 1.1   & 24    & 0.06  & 50 \\
    BA    & 1     & 0     & 0     & 2     & 0     & 1     & M     & N     & A     & 24    & 11    & 24    & 0.26  & 24    & 1.6   & 24    & 0.08  & 80 \\
    BE    & 1     & 1     & 0     & 1     & 0     & 0     & M     & A     & A     & 24    & 14    & 24    & 0.14  & 24    & 0.8   & 24    & 0.04  & 50 \\
    BG    & 1     & 0     & 0     & 1     & 0     & 0     & M     & N     & M     & 11    & 11    & 11    & 0.18  & 11    & 0.9   & 11    & 0.06  & 100 \\
    CH    & 2     & 1     & 4     & 0     & 0     & 2     & M     & A     & A     & 24    & 22    & 24    & 0.18  & 24    & 1.1   & 24    & 0.06  & 180 \\
    CY    & 1     & 0     & 0     & 1     & 1     & 0     & M     & N     & A     & 24    & 4     & 24    & 0.28  & 24    & 1.6   & 24    & 0.08  & 180 \\
    CZ    & 1     & 1     & 3     & 0     & 0     & 2     & M     & N     & A     & 15    & 16    & 16    & 0.16  & 16    & 0.8   & 16    & 0.04  & 140 \\
    DE    & 2     & 1     & 0     & 1     & 0     & 0     & M     & N     & M     & 19    & 11    & 18    & 0.18  & 21    & 1.05  & 18    & 0.04  & 180 \\
    DK    & 0     & 0     & 0     & 1     & 1     & 0     & A     & N     & A     & 10    & 7     & 10    & 0.38  & 10    & 1.4   & 10    & 0.16  & 190 \\
    EE    & 1     & 0     & 0     & 1     & 0     & 0     & M     & N     & M     & 12    & 3     & 18    & 0.2   & 17    & 1     & 22    & 0.06  & 190 \\
    ES    & 1     & 1     & 2     & 1     & 0     & 0     & M     & A     & M     & 24    & 21    & 24    & 0.3   & 24    & 1.8   & 24    & 0.08  & 190 \\
    FI    & 0     & 0     & 0     & 1     & 1     & 0     & M     & N     & M     & 22    & 2     & 23    & 0.14  & 23    & 0.7   & 24    & 0.04  & 50 \\
    FR    & 1     & 1     & 0     & 1     & 0     & 0     & M     & A     & M     & 13    & 21    & 12    & 0.14  & 12    & 0.7   & 15    & 0.06  & 200 \\
    GB    & 0     & 0     & 0     & 1     & 1     & 0     & M     & N     & M     & 13    & 6     & 12    & 0.26  & 12    & 1.2   & 12    & 0.1   & 7 \\
    GR    & 1     & 1     & 1     & 0     & 1     & 2     & M     & A     & M     & 24    & 21    & 24    & 0.28  & 24    & 1.65  & 24    & 0.08  & 120 \\
    HR    & 2     & 1     & 1     & 1     & 1     & 1     & A     & Ad    & A     & 12    & 13    & 12    & 0.2   & 12    & 1.05  & 12    & 0.06  & 170 \\
    HU    & 1     & 1     & 0     & 1     & 0     & 0     & A     & N     & A     & 24    & 8     & 24    & 0.2   & 24    & 1.2   & 24    & 0.06  & 190 \\
    IE    & 0     & 1     & 1     & 0     & 1     & 0     & M     & N     & M     & 13    & 4     & 24    & 0.22  & 16    & 1.2   & 24    & 0.06  & 35 \\
    IS    & 0     & 1     & 1     & 1     & 0     & 0     & M     & N     & A     & 18    & 5     & 18    & 0.26  & 17    & 1.2   & 18    & 0.04  & 180 \\
    IT    & 1     & 1     & 1     & 0     & 1     & 1     & M     & A     & M     & 24    & 14    & 13    & 0.26  & 12    & 1.25  & 24    & 0.06  & 1 \\
    LT    & 1     & 0     & 0     & 1     & 0     & 0     & M     & N     & A     & 24    & 2     & 24    & 0.18  & 24    & 1.1   & 24    & 0.04  & 140 \\
    LU    & 3     & 1     & 1     & 1     & 0     & 0     & M     & N     & A     & 22    & 20    & 24    & 0.28  & 24    & 1.6   & 24    & 0.06  & 180 \\
    LV    & 3     & 0     & 2     & 1     & 0     & 0     & M     & N     & A     & 23    & 4     & 23    & 0.22  & 23    & 1.25  & 23    & 0.06  & 110 \\
    ME    & 1     & 0     & 1     & 0     & 1     & 0     & M     & N     & M     & 22    & 10    & 9     & 0.4   & 9     & 1.6   & 9     & 0.14  & 25 \\
    MK    & 1     & 0     & 1     & 2     & 1     & 0     & M     & N     & A     & 12    & 7     & 11    & 0.14  & 11    & 0.65  & 12    & 0.04  & 140 \\
    NI    & 1     & 0     & 0     & 1     & 0     & 0     & M     & N     & A     & 12    & 21    & 6     & 0.02  & 10    & 1.95  & 10    & 0.2   & 140 \\
    NL    & 2     & 1     & 1     & 0     & 1     & 1     & M     & A     & M     & 23    & 33    & 23    & 0.28  & 24    & 1.7   & 18    & 0.08  & 190 \\
    NO    & 1     & 0     & 0     & 1     & 0     & 0     & M     & N     & M     & 16    & 3     & 17    & 0.16  & 16    & 0.65  & 16    & 0.04  & 160 \\
    PL    & 0     & 1     & 2     & 0     & 0     & 2     & M     & N     & A     & 19    & 15    & 18    & 0.22  & 22    & 1.3   & 17    & 0.06  & 80 \\
    PT    & 1     & 1     & 2     & 0     & 1     & 1     & M     & A     & M     & 24    & 17    & 24    & 0.24  & 24    & 1.5   & 24    & 0.08  & 170 \\
    RO    & 1     & 0     & 1     & 1     & 1     & 1     & M     & N     & A     & 24    & 14    & 18    & 0.26  & 18    & 1.4   & 19    & 0.06  & 45 \\
    RS    & 0     & 0     & 1     & 0     & 1     & 1     & M     & N     & M     & 11    & 16    & 11    & 0.36  & 11    & 1.65  & 10    & 0.12  & 35 \\
    SE    & 2     & 0     & 1     & 1     & 0     & 0     & M     & N     & M     & 12    & 3     & 12    & 0.1   & 12    & 0.45  & 12    & 0.02  & 20 \\
    SI    & 2     & 1     & 1     & 1     & 0     & 0     & A     & N     & A     & 24    & 28    & 24    & 0.32  & 24    & 1.95  & 24    & 0.08  & 200 \\
    SK    & 1     & 0     & 0     & 0     & 0     & 2     & M     & N     & A     & 17    & 12    & 17    & 0.14  & 24    & 1     & 18    & 0.04  & 150 \\
\bottomrule
    \end{tabular}%
\end{table}%

\begin{table}[]
\small
  \centering
      \setlength{\tabcolsep}{6pt}
  \caption{Hyperparameters of ETS for forecasting mean yearly load and yearly dispersion in "+ETS" baseline models. Here, TS is the time series ID; error, trend and seasonal denote components of ETS; and M, N and A refer to whether the component of ETS is multiplicative, none or additive, respectively.}
   \label{tabA2}%
    \begin{tabular}{c|ccc|ccc}
    
\toprule
    \multicolumn{1}{c|}{TS}  & \multicolumn{3}{c|}{ETS mean} & \multicolumn{3}{c}{ETS dispersion} \\
           & error & trend & seasonal & error & trend & seasonal \\
\midrule
   
   AT    & M     & A     & N     & A     & N     & N \\
    BA    & A     & N     & N     & M     & N     & N \\
    BE    & M     & A     & N     & A     & N     & N \\
    BG    & M     & N     & N     & A     & N     & N \\
    CH    & A     & A     & N     & A     & N     & N \\
    CY    & A     & A     & N     & A     & N     & N \\
    CZ    & M     & N     & N     & A     & A     & N \\
    DE    & A     & N     & N     & M     & A     & N \\
    DK    & A     & A     & N     & A     & A     & N \\
    EE    & A     & N     & N     & A     & N     & N \\
    ES    & M     & A     & N     & A     & N     & N \\
    FI    & A     & N     & N     & A     & N     & N \\
    FR    & M     & A     & N     & M     & N     & N \\
    GB    & A     & N     & N     & A     & N     & N \\
    GR    & M     & A     & N     & M     & A     & N \\
    HR    & M     & A     & N     & M     & N     & N \\
    HU    & M     & N     & N     & A     & N     & N \\
    IE    & A     & N     & N     & A     & N     & N \\
    IS    & A     & N     & N     & A     & N     & N \\
    IT    & M     & A     & N     & M     & N     & N \\
    LT    & A     & N     & N     & A     & N     & N \\
    LU    & M     & A     & N     & A     & N     & N \\
    LV    & A     & N     & N     & A     & N     & N \\
    ME    & A     & N     & N     & A     & A     & N \\
    MK    & M     & N     & N     & M     & N     & N \\
    NI    & A     & A     & N     & A     & N     & N \\
    NL    & M     & A     & N     & A     & A     & N \\
    NO    & A     & N     & N     & A     & A     & N \\
    PL    & A     & N     & N     & A     & N     & N \\
    PT    & M     & A     & N     & M     & N     & N \\
    RO    & M     & N     & N     & M     & N     & N \\
    RS    & A     & N     & N     & A     & N     & N \\
    SE    & A     & N     & N     & A     & A     & N \\
    SI    & A     & N     & N     & M     & N     & N \\
    SK    & M     & N     & N     & A     & A     & N \\

\bottomrule
   
    \end{tabular}%
 
\end{table}%

\begin{table}[]
\small
  \centering
    \setlength{\tabcolsep}{2pt}
  \caption{Hyperparameters for MLP. Here, TS is the time series ID, $n$ is input pattern length and $h$ is forecast horizon.}
  \label{tabA3}%
    \begin{tabular}{c|cccccccccccc|cccccccccccc}
\toprule
\multicolumn{1}{c|}{TS} & \multicolumn{12}{c|}{$n$}                                                                       & \multicolumn{12}{c}{\#hidden nodes} \\
          & \multicolumn{1}{c}{$h$=1} & 2     & 3     & 4     & 5     & 6     & 7     & 8     & 9     & 10    & 11    & 12    & $h$=1     & 2     & 3     & 4     & 5     & 6     & 7     & 8     & 9     & 10    & 11    & 12 \\
          \midrule
    AT    & 24    & 24    & 24    & 24    & 24    & 24    & 24    & 24    & 24    & 12    & 24    & 23    & 9     & 1     & 5     & 1     & 9     & 4     & 4     & 9     & 6     & 4     & 3     & 1 \\
    BA    & 22    & 21    & 24    & 13    & 24    & 17    & 23    & 24    & 19    & 20    & 24    & 12    & 2     & 4     & 5     & 7     & 5     & 8     & 1     & 1     & 8     & 6     & 1     & 9 \\
    BE    & 12    & 24    & 12    & 12    & 24    & 24    & 15    & 24    & 18    & 24    & 19    & 10    & 9     & 2     & 8     & 2     & 1     & 10    & 3     & 2     & 6     & 9     & 7     & 2 \\
    BG    & 24    & 20    & 23    & 23    & 12    & 5     & 6     & 17    & 23    & 24    & 23    & 23    & 8     & 9     & 8     & 7     & 6     & 1     & 10    & 9     & 4     & 10    & 10    & 2 \\
    CH    & 12    & 15    & 12    & 12    & 12    & 16    & 12    & 12    & 12    & 20    & 21    & 13    & 1     & 3     & 10    & 9     & 8     & 1     & 5     & 2     & 4     & 3     & 1     & 9 \\
    CY    & 11    & 9     & 5     & 3     & 5     & 12    & 4     & 6     & 7     & 6     & 12    & 4     & 4     & 1     & 10    & 9     & 9     & 2     & 5     & 6     & 5     & 8     & 4     & 8 \\
    CZ    & 24    & 23    & 24    & 24    & 12    & 12    & 17    & 18    & 19    & 21    & 12    & 12    & 1     & 5     & 1     & 7     & 4     & 10    & 10    & 10    & 4     & 3     & 8     & 10 \\
    DE    & 12    & 23    & 12    & 24    & 24    & 12    & 12    & 12    & 8     & 12    & 11    & 12    & 5     & 8     & 10    & 5     & 7     & 1     & 2     & 2     & 10    & 4     & 2     & 10 \\
    DK    & 11    & 7     & 12    & 10    & 11    & 9     & 9     & 12    & 11    & 11    & 11    & 12    & 9     & 5     & 7     & 6     & 5     & 3     & 4     & 7     & 1     & 8     & 1     & 4 \\
    EE    & 11    & 9     & 10    & 11    & 12    & 12    & 12    & 12    & 12    & 9     & 10    & 11    & 4     & 5     & 5     & 9     & 2     & 4     & 6     & 9     & 7     & 2     & 8     & 3 \\
    ES    & 24    & 12    & 12    & 24    & 12    & 12    & 12    & 12    & 12    & 6     & 24    & 19    & 9     & 5     & 1     & 8     & 10    & 6     & 4     & 2     & 6     & 7     & 7     & 2 \\
    FI    & 12    & 3     & 12    & 10    & 12    & 12    & 12    & 7     & 8     & 10    & 10    & 11    & 10    & 9     & 4     & 1     & 1     & 6     & 2     & 3     & 6     & 9     & 6     & 9 \\
    FR    & 12    & 7     & 12    & 12    & 12    & 12    & 12    & 18    & 12    & 24    & 12    & 23    & 1     & 5     & 7     & 10    & 1     & 8     & 3     & 9     & 2     & 1     & 4     & 1 \\
    GB    & 3     & 5     & 3     & 8     & 11    & 12    & 12    & 12    & 12    & 11    & 10    & 12    & 8     & 7     & 8     & 3     & 8     & 6     & 6     & 3     & 9     & 6     & 3     & 7 \\
    GR    & 23    & 12    & 12    & 12    & 8     & 12    & 12    & 12    & 12    & 12    & 8     & 8     & 9     & 8     & 6     & 6     & 8     & 4     & 2     & 3     & 3     & 7     & 10    & 6 \\
    HR    & 24    & 12    & 14    & 11    & 12    & 5     & 17    & 23    & 5     & 18    & 18    & 12    & 10    & 6     & 2     & 5     & 7     & 5     & 10    & 6     & 10    & 6     & 10    & 1 \\
    HU    & 24    & 23    & 15    & 24    & 24    & 12    & 22    & 12    & 6     & 12    & 12    & 15    & 8     & 1     & 7     & 2     & 7     & 3     & 6     & 1     & 6     & 3     & 4     & 4 \\
    IE    & 11    & 9     & 5     & 10    & 4     & 12    & 6     & 12    & 12    & 10    & 10    & 11    & 5     & 3     & 4     & 7     & 10    & 5     & 1     & 6     & 1     & 6     & 7     & 1 \\
    IS    & 4     & 8     & 3     & 3     & 3     & 12    & 12    & 12    & 4     & 8     & 8     & 11    & 5     & 4     & 7     & 8     & 8     & 1     & 8     & 7     & 3     & 3     & 10    & 3 \\
    IT    & 12    & 19    & 10    & 12    & 24    & 12    & 21    & 24    & 12    & 12    & 21    & 23    & 10    & 10    & 1     & 9     & 7     & 7     & 1     & 1     & 8     & 3     & 2     & 7 \\
    LT    & 12    & 9     & 12    & 12    & 11    & 11    & 5     & 6     & 7     & 8     & 12    & 3     & 6     & 3     & 4     & 7     & 9     & 1     & 3     & 8     & 3     & 10    & 10    & 7 \\
    LU    & 24    & 12    & 12    & 12    & 12    & 12    & 24    & 24    & 18    & 21    & 4     & 3     & 9     & 6     & 2     & 9     & 5     & 9     & 9     & 8     & 5     & 3     & 8     & 2 \\
    LV    & 7     & 3     & 12    & 12    & 11    & 5     & 12    & 7     & 7     & 12    & 10    & 12    & 7     & 4     & 7     & 7     & 5     & 9     & 9     & 4     & 9     & 10    & 8     & 10 \\
    ME    & 12    & 10    & 6     & 8     & 7     & 10    & 11    & 11    & 10    & 11    & 11    & 18    & 6     & 9     & 6     & 2     & 5     & 3     & 6     & 1     & 6     & 10    & 2     & 8 \\
    MK    & 12    & 12    & 12    & 24    & 13    & 5     & 6     & 12    & 6     & 14    & 12    & 12    & 5     & 3     & 1     & 9     & 9     & 6     & 7     & 2     & 2     & 2     & 5     & 8 \\
    NI    & 11    & 11    & 11    & 3     & 11    & 9     & 6     & 12    & 8     & 12    & 11    & 11    & 5     & 3     & 6     & 5     & 3     & 9     & 1     & 1     & 5     & 1     & 7     & 2 \\
    NL    & 6     & 23    & 12    & 12    & 12    & 12    & 12    & 12    & 7     & 12    & 8     & 10    & 2     & 5     & 2     & 3     & 1     & 8     & 4     & 1     & 7     & 5     & 8     & 2 \\
    NO    & 12    & 11    & 12    & 3     & 12    & 12    & 12    & 12    & 8     & 9     & 10    & 10    & 6     & 2     & 5     & 6     & 8     & 3     & 7     & 4     & 9     & 2     & 1     & 7 \\
    PL    & 12    & 12    & 12    & 24    & 24    & 24    & 24    & 22    & 24    & 10    & 12    & 24    & 1     & 3     & 2     & 6     & 4     & 1     & 9     & 3     & 6     & 4     & 3     & 10 \\
    PT    & 24    & 12    & 12    & 12    & 12    & 12    & 12    & 12    & 12    & 12    & 12    & 12    & 9     & 2     & 3     & 9     & 10    & 9     & 1     & 6     & 4     & 10    & 7     & 7 \\
    RO    & 12    & 12    & 24    & 24    & 12    & 12    & 12    & 11    & 7     & 18    & 12    & 24    & 7     & 6     & 5     & 8     & 2     & 4     & 5     & 4     & 8     & 6     & 5     & 4 \\
    RS    & 23    & 5     & 23    & 18    & 11    & 9     & 24    & 11    & 8     & 24    & 13    & 16    & 8     & 7     & 6     & 6     & 2     & 2     & 3     & 5     & 8     & 8     & 1     & 9 \\
    SE    & 3     & 3     & 12    & 9     & 12    & 12    & 12    & 12    & 8     & 9     & 12    & 11    & 1     & 10    & 8     & 4     & 7     & 1     & 2     & 7     & 2     & 6     & 7     & 4 \\
    SI    & 24    & 17    & 24    & 23    & 11    & 19    & 12    & 12    & 19    & 22    & 12    & 12    & 8     & 3     & 6     & 1     & 10    & 3     & 8     & 9     & 3     & 5     & 1     & 5 \\
    SK    & 24    & 12    & 12    & 12    & 12    & 24    & 12    & 18    & 20    & 24    & 12    & 13    & 1     & 1     & 1     & 1     & 7     & 3     & 4     & 1     & 9     & 7     & 1     & 5 \\
\bottomrule
   \end{tabular}%
\end{table}%

\begin{table}[]
\small
  \centering
      \setlength{\tabcolsep}{2pt}
  \caption{Hyperparameters for ANFIS. Here, TS is the time series ID, $n$ is input pattern length and $h$ is forecast horizon.}
   \label{tabA4}%
    \begin{tabular}{c|cccccccccccc|cccccccccccc}
\toprule
\multicolumn{1}{c|}{TS} & \multicolumn{12}{c|}{$n$}                                                                       & \multicolumn{12}{c}{\#rules} \\
          & \multicolumn{1}{c}{$h$=1} & 2     & 3     & 4     & 5     & 6     & 7     & 8     & 9     & 10    & 11    & 12    & $h$=1     & 2     & 3     & 4     & 5     & 6     & 7     & 8     & 9     & 10    & 11    & 12 \\
\midrule
    AT    & 24    & 23    & 13    & 23    & 12    & 24    & 24    & 12    & 19    & 24    & 24    & 13    & 6     & 8     & 10    & 10    & 13    & 6     & 6     & 13    & 8     & 7     & 5     & 7 \\
    BA    & 24    & 23    & 22    & 24    & 23    & 13    & 12    & 12    & 20    & 12    & 19    & 24    & 3     & 4     & 6     & 5     & 3     & 3     & 2     & 2     & 3     & 4     & 5     & 2 \\
    BE    & 12    & 24    & 24    & 12    & 24    & 24    & 24    & 18    & 24    & 24    & 23    & 21    & 2     & 4     & 8     & 7     & 8     & 7     & 9     & 10    & 2     & 12    & 11    & 12 \\
    BG    & 13    & 12    & 8     & 23    & 22    & 22    & 10    & 12    & 23    & 22    & 23    & 24    & 5     & 2     & 7     & 5     & 2     & 5     & 2     & 2     & 7     & 5     & 4     & 2 \\
    CH    & 23    & 12    & 3     & 23    & 14    & 14    & 12    & 12    & 24    & 20    & 23    & 17    & 11    & 10    & 1     & 4     & 5     & 6     & 13    & 13    & 9     & 10    & 13    & 10 \\
    CY    & 12    & 5     & 5     & 5     & 5     & 12    & 9     & 9     & 12    & 5     & 5     & 5     & 1     & 1     & 9     & 1     & 1     & 1     & 1     & 1     & 1     & 1     & 1     & 1 \\
    CZ    & 22    & 10    & 22    & 24    & 24    & 24    & 24    & 12    & 19    & 24    & 24    & 10    & 12    & 6     & 6     & 9     & 6     & 11    & 7     & 10    & 11    & 9     & 4     & 5 \\
    DE    & 15    & 9     & 24    & 11    & 24    & 12    & 12    & 24    & 24    & 10    & 21    & 12    & 4     & 5     & 4     & 10    & 2     & 12    & 6     & 3     & 5     & 3     & 12    & 13 \\
    DK    & 3     & 5     & 4     & 9     & 9     & 8     & 12    & 9     & 10    & 11    & 5     & 4     & 1     & 1     & 1     & 1     & 1     & 6     & 2     & 1     & 11    & 1     & 1     & 1 \\
    EE    & 3     & 4     & 4     & 7     & 9     & 10    & 8     & 10    & 9     & 8     & 6     & 4     & 1     & 1     & 1     & 1     & 1     & 5     & 3     & 2     & 7     & 2     & 1     & 3 \\
    ES    & 11    & 24    & 24    & 12    & 24    & 24    & 12    & 11    & 24    & 12    & 8     & 24    & 6     & 10    & 8     & 10    & 12    & 3     & 6     & 1     & 3     & 5     & 4     & 3 \\
    FI    & 3     & 3     & 4     & 10    & 9     & 9     & 9     & 9     & 9     & 7     & 6     & 4     & 1     & 1     & 1     & 1     & 1     & 1     & 1     & 1     & 1     & 1     & 1     & 1 \\
    FR    & 24    & 23    & 15    & 12    & 12    & 12    & 24    & 24    & 24    & 12    & 24    & 23    & 13    & 13    & 5     & 11    & 11    & 5     & 7     & 12    & 9     & 7     & 4     & 7 \\
    GB    & 3     & 5     & 4     & 10    & 9     & 9     & 9     & 11    & 9     & 9     & 5     & 4     & 1     & 1     & 1     & 1     & 1     & 1     & 6     & 8     & 1     & 1     & 1     & 1 \\
    GR    & 20    & 12    & 23    & 19    & 24    & 23    & 21    & 24    & 6     & 24    & 12    & 12    & 4     & 3     & 12    & 2     & 10    & 10    & 5     & 13    & 1     & 8     & 4     & 3 \\
    HR    & 12    & 12    & 14    & 23    & 24    & 12    & 24    & 14    & 24    & 8     & 23    & 10    & 4     & 7     & 2     & 9     & 3     & 7     & 8     & 1     & 6     & 11    & 3     & 8 \\
    HU    & 12    & 24    & 15    & 24    & 24    & 11    & 24    & 24    & 7     & 12    & 3     & 3     & 11    & 4     & 5     & 5     & 5     & 3     & 1     & 10    & 3     & 6     & 1     & 10 \\
    IE    & 3     & 6     & 5     & 10    & 9     & 9     & 9     & 11    & 9     & 7     & 5     & 3     & 1     & 1     & 1     & 1     & 1     & 1     & 1     & 11    & 1     & 1     & 1     & 1 \\
    IS    & 5     & 9     & 5     & 4     & 3     & 9     & 4     & 4     & 9     & 3     & 3     & 5     & 1     & 1     & 1     & 1     & 1     & 1     & 1     & 1     & 1     & 1     & 1     & 1 \\
    IT    & 22    & 24    & 24    & 24    & 24    & 23    & 24    & 24    & 12    & 24    & 24    & 22    & 3     & 9     & 9     & 4     & 3     & 8     & 4     & 6     & 2     & 3     & 7     & 2 \\
    LT    & 3     & 3     & 3     & 12    & 9     & 9     & 9     & 9     & 8     & 4     & 3     & 3     & 1     & 1     & 1     & 11    & 1     & 1     & 1     & 1     & 2     & 1     & 1     & 1 \\
    LU    & 22    & 24    & 23    & 12    & 24    & 24    & 24    & 24    & 12    & 7     & 12    & 12    & 9     & 9     & 9     & 10    & 8     & 10    & 10    & 4     & 1     & 13    & 13    & 12 \\
    LV    & 3     & 4     & 3     & 9     & 11    & 11    & 9     & 11    & 9     & 8     & 5     & 3     & 1     & 1     & 1     & 1     & 7     & 6     & 4     & 9     & 6     & 2     & 1     & 1 \\
    ME    & 12    & 3     & 4     & 9     & 7     & 8     & 5     & 12    & 12    & 13    & 8     & 15    & 4     & 1     & 1     & 3     & 4     & 3     & 4     & 2     & 2     & 2     & 4     & 2 \\
    MK    & 14    & 12    & 12    & 8     & 17    & 17    & 24    & 19    & 22    & 9     & 11    & 15    & 2     & 5     & 7     & 11    & 3     & 3     & 6     & 3     & 6     & 7     & 7     & 6 \\
    NI    & 3     & 6     & 4     & 10    & 9     & 12    & 12    & 9     & 9     & 6     & 5     & 3     & 1     & 1     & 1     & 1     & 1     & 11    & 8     & 9     & 1     & 1     & 1     & 1 \\
    NL    & 7     & 5     & 12    & 22    & 24    & 12    & 12    & 24    & 23    & 24    & 24    & 10    & 11    & 1     & 10    & 12    & 6     & 13    & 12    & 9     & 13    & 11    & 7     & 6 \\
    NO    & 3     & 3     & 4     & 6     & 9     & 9     & 11    & 9     & 8     & 7     & 5     & 3     & 1     & 1     & 1     & 10    & 4     & 1     & 13    & 2     & 2     & 1     & 1     & 1 \\
    PL    & 12    & 23    & 11    & 12    & 11    & 24    & 24    & 18    & 23    & 9     & 10    & 12    & 6     & 5     & 13    & 2     & 2     & 3     & 2     & 9     & 5     & 11    & 6     & 2 \\
    PT    & 22    & 4     & 11    & 11    & 12    & 12    & 24    & 12    & 12    & 12    & 12    & 12    & 4     & 1     & 13    & 3     & 12    & 13    & 13    & 7     & 3     & 5     & 3     & 4 \\
    RO    & 12    & 20    & 12    & 12    & 11    & 18    & 19    & 16    & 12    & 19    & 9     & 22    & 3     & 6     & 4     & 5     & 2     & 2     & 2     & 7     & 4     & 2     & 8     & 4 \\
    RS    & 9     & 5     & 19    & 10    & 21    & 23    & 24    & 11    & 12    & 12    & 23    & 22    & 3     & 4     & 2     & 1     & 2     & 2     & 3     & 2     & 4     & 4     & 1     & 2 \\
    SE    & 3     & 4     & 4     & 8     & 9     & 9     & 11    & 7     & 8     & 9     & 6     & 4     & 1     & 1     & 1     & 1     & 1     & 1     & 13    & 8     & 6     & 1     & 1     & 1 \\
    SI    & 14    & 24    & 23    & 23    & 21    & 10    & 12    & 23    & 23    & 12    & 18    & 16    & 11    & 1     & 11    & 13    & 4     & 4     & 1     & 11    & 1     & 5     & 9     & 11 \\
    SK    & 17    & 20    & 24    & 15    & 12    & 23    & 24    & 24    & 18    & 21    & 11    & 23    & 2     & 6     & 10    & 4     & 2     & 10    & 12    & 9     & 8     & 8     & 13    & 9 \\
\bottomrule
   \end{tabular}%

\end{table}%

\end{document}